\documentclass[10pt,twocolumn,letterpaper]{article}

\usepackage[pagenumbers]{cvpr} 

\usepackage{graphicx}
\usepackage{amsmath}
\usepackage{amssymb}
\usepackage{booktabs}
\usepackage{bm}
\usepackage[utf8]{inputenc}
\usepackage[margin=2.5cm]{geometry}
\usepackage{enumitem}
\usepackage{multirow}
\usepackage[accsupp]{axessibility}

\definecolor{cvprblue}{rgb}{0.21,0.49,0.74}
\usepackage[pagebackref,breaklinks,colorlinks,allcolors=cvprblue]{hyperref}

\newcommand{\myparagraph}[1]{\vspace{2pt}\noindent\textbf{#1}.}
\def\paperID{0019} 
\def\confName{CVPR}
\def\confYear{2025}

\title{Video, How Do Your Tokens Merge?}

\author{Sam Pollard\\
University of Bristol\\
{\tt\small sam.pollard@bristol.ac.uk}\\
\and
Michael Wray\\
University of Bristol\\
{\tt\small michael.wray@bristol.ac.uk}\\
}

\begin{document}
\twocolumn[{
\maketitle
\begin{center}
    \captionsetup{type=figure}
    \includegraphics[width=\textwidth]{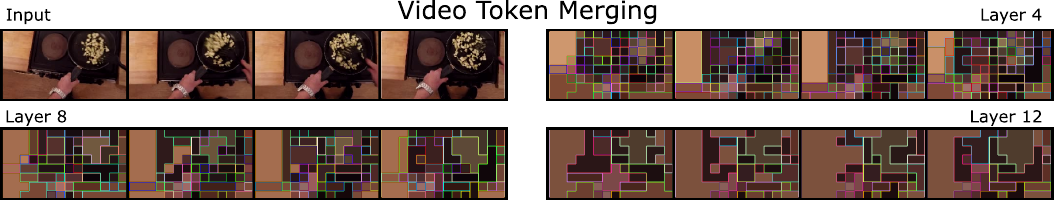}
    \captionof{figure}{Video Token Merging reduces computation of video transformer models by successively merging tokens without re-training or additional learned parameters. We show how an input video has its tokens merged across different layers.}
\end{center}
}]
\begin{abstract}
   Video transformer models require huge amounts of compute resources due to the spatio-temporal scaling of the input.
   Tackling this, recent methods have proposed to drop or merge tokens for image models, whether randomly or via learned methods.
   Merging tokens has many benefits: it can be plugged into any vision transformer, does not require model re-training, and it propagates information that would otherwise be dropped through the model.
   Before now, video token merging has not been evaluated on temporally complex datasets for video understanding.
   In this work, we explore training-free token merging for video to provide comprehensive experiments and find best practices across four video transformers on three datasets that exhibit coarse and fine-grained action recognition.
   Our results showcase the benefits of video token merging with a speedup of around $2.5$X while maintaining accuracy (avg. $-0.55\%$ for ViViT).
   Code available at~\href{https://github.com/sjpollard/video-how-do-your-tokens-merge}{https://github.com/sjpollard/video-how-do-your-tokens-merge}.
\end{abstract}    
\section{Introduction}
\label{sec:intro}

Vision transformers~\cite{dosovitskiy2020image} (ViTs) have quickly been established as the architecture of choice for solving the majority of computer vision problems. 
The long range dependencies that are afforded by self-attention significantly improve the flexibility and reasoning of models, when supplied with enough training data. 
This performance is not free as the quadratic complexity of the attention mechanism leads to exceedingly poor scaling. 

This is especially prevalent in the video domain, in which the sequence length of tokens increases with both spatial and temporal dimensions.
In the recent push for long video understanding~\cite{soldan2022mad,carreira2024learning,grauman2022ego4d}, video vision transformers are therefore requiring more and more resources for both training and inference.
Assuming that a $5$ minute video at $224 \times 224$ pixels consists of the typical $16 \times 16$ spatial patches, when sampling $1$ frame per second, the transformer sequence length is $58,800$ tokens.
With this increase in complexity, it's simply not enough to hope that hardware scales up at the same rate: this requires a huge number of resources per video and is significantly longer than the sequence lengths that can be handled in a single window of context for current transformers.
It is also worth noting that this sampling rate may not even be enough for fine-grained action understanding, where actions average $2.6$s in duration~\cite{damen2022rescaling}.

\begin{figure*}[!htp]
    \centering
    \includegraphics[width=0.98\linewidth]{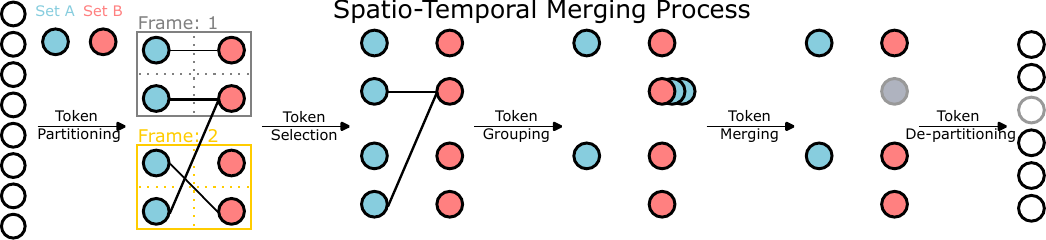}
    \caption{The merging process first separates tokens into two sets. Similarities are calculated and a one-to-many bipartite matching between tokens in each set is found. Finally, the top $r$ edges are kept and these are merged based on the similarity between tokens.}
    \label{fig:video_merging}
\end{figure*}

Recently, token merging of transformers has been proposed for images~\cite{bolya2022token}, as a drop-in method for increasing the throughput of ViT models by reducing the token sequence length.
This has several benefits over other methods: the method does not require re-training of a pre-trained model and the metric to merge tokens is already calculated as part of the forward pass.
In this work, \textit{we explore the implementation of token merging in the video domain without fine-tuning}, allowing tokens to merge freely in and between frames, to showcase its viability on video models that have been built upon ViT.
We evaluate video token merging on Kinetics-$400$~\cite{kay2017kinetics}, Something-Something v$2$~\cite{goyal2017something}, and EPIC-KITCHENS-$100$~\cite{damen2022rescaling}, which are datasets commonly used for video understanding, using both third and first person videos as well as containing actions with differing granularities.

To summarise, our contributions are as follows: (i) we explore best practices for training-free token merging for video; (ii) we provide a comprehensive evaluation of video transformer models on common action recognition benchmarks, demonstrating a training-free speedup of $2.5$X; and (iii) we provide an in-depth analysis of how tokens are merged within spatio-temporal models.

\section{Related Work}
\label{sec:related_work}

Efficiency of transformers is a popular topic, with a variety of possible approaches.
Recent works in natural language processing (NLP) have focused on the creation of more computationally efficient attention modules to reduce the overhead of the quadratic complexity~\cite{wang2020linformer}, attention modules that make better use of modern hardware~\cite{dao2022flashattention}, and new architectures that scale linearly with input size~\cite{gu2023mamba}.
Others make use of domain specific knowledge to reduce the range of tokens that self-attention is applied to in vision problems~\cite{bertasius2021space,patrick2021keeping}.
Alternatively, in~\cite{touvron2021training}, knowledge distillation is applied to make transformers more efficient learners at training time.
Transformers are resilient to the dropping (or pruning) of input tokens, attention heads, and even entire blocks, which~\cite{meng2022adavit} has demonstrated.
The dropout of input tokens is of particular interest for vision transformers because it exploits the data redundancy present in the natural image and video domain.
Accordingly, we now introduce works whose focus is to reduce the token sequence length, whether by dropping them directly or merging them into more compact features.

\subsection{Token Dropout}
\label{sub:token_dropout}

A simple method to reduce the sequence length in transformers is to drop tokens, i.e. token dropout. 
This has been applied to vision problems in multiple ways: tokens can be dropped at random~\cite{han2022turbo,liu2023patchdropout} or via a learned module~\cite{rao2021dynamicvit,yin2022vit,chen2023efficient}. 
In particular, \cite{liu2023patchdropout} introduces a scheme whereby image classification training is expedited by applying random token dropout to vanilla ViTs, however, at inference time the whole token sequence is used. 
Similarly, \cite{han2022turbo} trains a partial masked auto-encoder for video understanding by dropping out significant portions of the input and then partially reconstructing the video. 
Dropout is well established in the literature, but emphasis has not been placed on the speedup of transformers at inference time \emph{without} re-training.

\vspace{-2pt}
\subsection{Token Merging}
\label{sub:token_merging}

Tokens have been merged in a variety of ways to reduce the information lost by dropout. 
In~\cite{liang2022not} innattentive tokens are fused into a single ``background'' token, while others exploit semantics to improve features of interest~\cite{zeng2022not,zhou2023can}. Other works~\cite{lee2024video} optimise token merging through trainable parameters. 
Most recently, \cite{koner2024lookupvit} trains models with an extra stream of compressed tokens to reduce attention overhead. 
Alternatively, there are many works that make use of token merging, as established in~\cite{bolya2022token,bolya2023token}, where image tokens are merged via a weighted average by reusing attention keys as a similarity metric. 
The primary motivation of the method is that it can be applied to ViTs \textit{without re-training or additional learned modules}. 
This has been extended to VLMs~\cite{cao2023pumer,ren2023testa,choi2024vid,weng2024longvlm, zhong2024aim}, semantic segmentation~\cite{kienzle2024segformer++}, and video editing~\cite{li2024vidtome}. 
In~\cite{kim2024token}, the merging method is improved to bridge the gap between dropout and merging.
We note that prior works have not explored video token merging  on challenging video datasets.
\section{Method}
\label{sec:method}

Our implementation of spatio-temporal token merging for video transformers is an extension of image based token merging~\cite{bolya2022token}.
We first explain token merging for videos in~\cref{sub:spatio_temporal_token_merging} before describing how token merging can be applied to methods which explicitly incorporate attention across the temporal dimension in~\cref{sub:divided_space_time_token_merging}.

\subsection{Spatio-Temporal Token Merging}
\label{sub:spatio_temporal_token_merging}

The goal of token merging is to reduce the number of tokens within the model, which reduces the overhead of attention calculation between all pairs of tokens, increasing throughput of the method.
Tokens are merged instead of dropping them, so that complementary information is preserved but redundant information is reduced, with the aim of preserving accuracy of the model.
The merging process is applied within each layer in the transformer successively, to ensure that information is not lost too quickly.
We give an overview of the process in~\cref{fig:video_merging} which is explained in greater depth below.

In detail, at the $i^\text{th}$ layer of the transformer, we define the video token sequence as $x \in \mathbb{R}^{B \times S^V_i \times D}$, where $B$ is the batch size, $S^V_i$ is the current number of tokens in the video sequence for the $i^\text{th}$ layer, and $D$ is the channel depth of the features. 
We first \textit{partition the tokens} $x$, into two sets $\mathbb{A}$ and $\mathbb{B}$, both of size $S^V_i/2$.
The tokens are partitioned in an alternating manner, as is established in~\cite{bolya2022token}.
Next, we \textit{select the most similar token} in $\mathbb{B}$ for every token in $\mathbb{A}$, creating a one-to-many matching between the two sets as a bipartite graph $G$.
Note that these connections are drawn \textit{across all} frames.
The similarity is defined as the cosine distance between the keys (K) of two tokens from the Query-Key-Value self-attention and this is used as the edge weighting in $G$.

We define a hyperparameter for the number of tokens to merge as $r_i$, where $0 \leq r_i \leq S^V_i/2$, which represents the number of tokens to merge within the $i\text{th}$ layer.
Accordingly, $r_i$ edges with the highest similarity (the edge weight) are kept and all other edges are dropped.
The remaining edges denote \textit{tokens to be merged} and their features are then averaged, weighted by their size.
The size of each token is tracked as $n \in \mathbb{R}^{S^V_i}$, a vector where each value represents the total number of tokens that have been merged into each location in the sequence.

After the \textit{tokens are de-partitioned}, each token in the sequence is now the weighted average of an arbitrary number of image patches. 
This has been demonstrated by~\cite{bolya2022token} to affect the attention output of the next layer; when tokens with similar keys are merged, they'll have a smaller impact on the softmax output.
To help alleviate this problem, \textit{proportional attention} is introduced:
\begin{equation}
    \text{softmax} \big(\frac{\bm{Q}\bm{K}^T}{\sqrt{d}} + \log n\big)
\end{equation}

ViViT~\cite{arnab2021vivit} and VideoMAE~\cite{tong2022videomae} extend the ViT architecture~\cite{dosovitskiy2020image} to incorporate the temporal dimension by jointly attending across space and time.
There are few architectural differences between these video models and image ViTs, with the additional temporal dimension making them computationally expensive.
To reduce this overhead, the patch embedding is 3D instead of 2D, meaning each token spans multiple frames.

\subsection{Divided Space-Time Token Merging}
\label{sub:divided_space_time_token_merging}

The works of TimeSformer~\cite{bertasius2021space} and Motionformer~\cite{patrick2021keeping} forgo joint space-time attention and instead experiment with methods to fuse information in a more efficient manner.
However, this requires the token sequence to have temporal structure, as the temporal attentions are calculated through frames, using the spatial position as an anchor point.
Because of this, merging tokens between frames causes temporal relationships to be broken and so we apply the token merging operation to each frame independently, as if they are unrelated images.
Note, that these tokens still encode temporal information through the attention process.
Accordingly, the hidden states are $x \in \mathbb{R}^{BF \times S^I_i \times D}$, where $B$ is the batch size, $F$ is the number of frames in the video, $S^I_i$ is the current number of tokens in the image sequence for the $i^\text{th}$ layer.
To the best of our knowledge, we are the first work to implement token merging in this manner with divided space-time transformers.

\begin{figure*}[!ht]
  \centering
  \begin{subfigure}{0.49\linewidth}
    \centering
    \includegraphics[width=\linewidth]{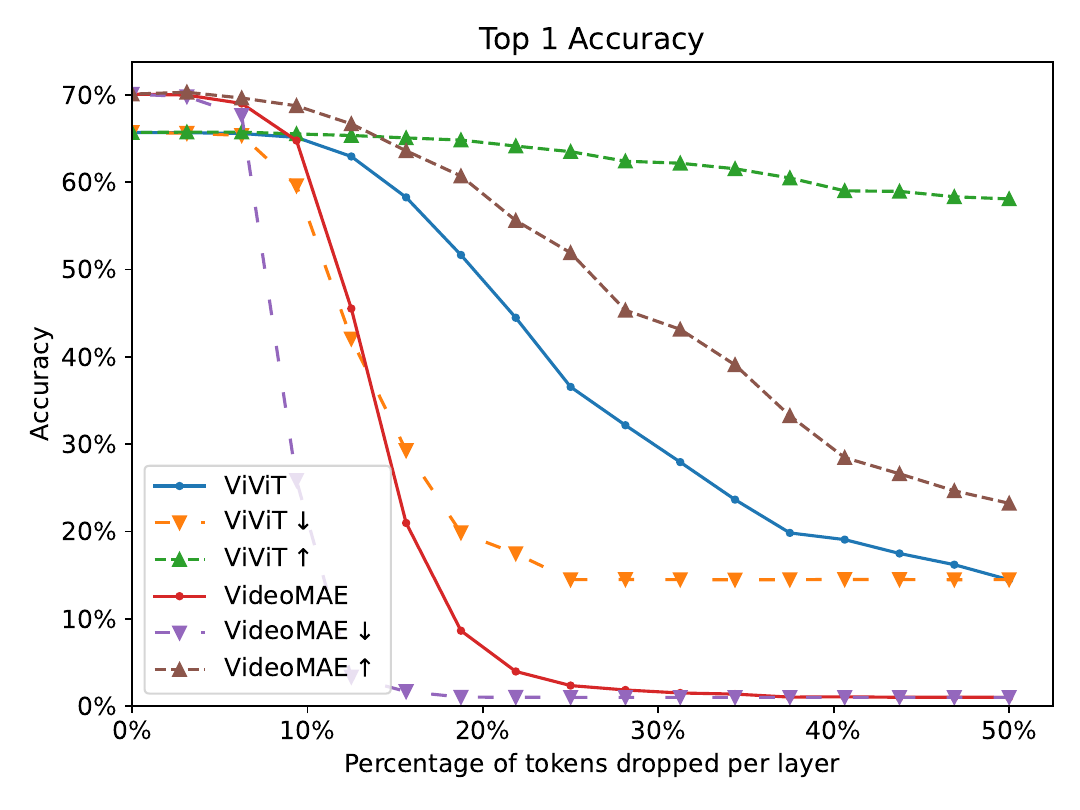}
    \caption{Accuracy as $r$ increases.}
    \label{fig:scaling_curves_a}
  \end{subfigure}
  \hfill
  \begin{subfigure}{0.49\linewidth}
    \centering
    \includegraphics[width=\linewidth]{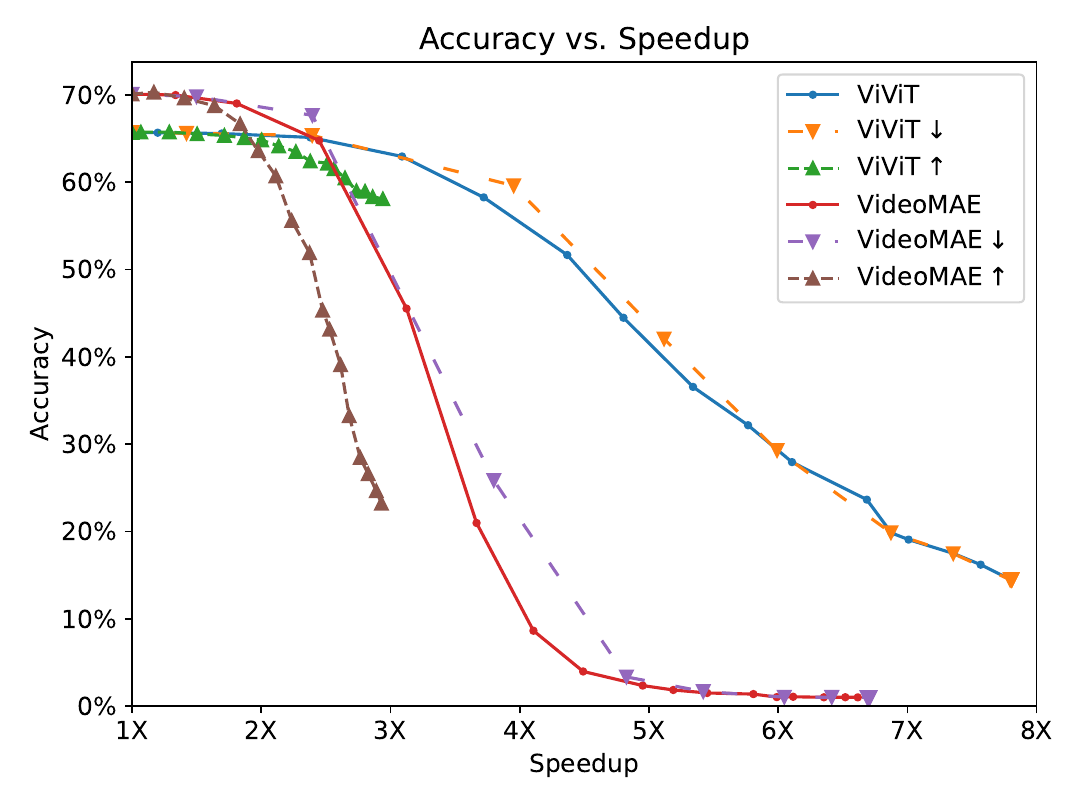}
    \caption{Accuracy against speedup}
    \label{fig:scaling_curves_b}
  \end{subfigure}
  \caption{(Left) curves corresponds to accuracy with ViViT and VideoMAE on K$400$ when increasing $r$ (the number of tokens merged) up to its limit. The $x$-axis is the percentage (relative to the original total) of tokens dropped \textit{per layer}. (Right) figure displays the accuracy against speedup gained for these $r$ values.}
  \label{fig:scaling_curves}
\end{figure*}
\section{Results}
\label{sec:results}

Within this section we present comprehensive experiments of video token merging on different video transformer methods across different datasets.
We aim to provide a consistent view of these methods and how the accuracy/speedup trade-off of each method can vary depending on the granularity of actions within each dataset.

\subsection{Implementation Details}
\label{sub:implementation_details}

We test token merging in video across a selection of four different transformer models. 
The models are as follows: TimeSformer with $8$ frames, Motionformer with $16$ frames, VideoMAE with $16$ frames and ViViT (``Model $1$'' from~\cite{arnab2021vivit}) with $32$ frames. 
In all of our experiments we use the base size models with default number of frames.
This has the benefit of being a fair testbed whilst also evaluating token merging performance across a variable number of frames. 
Where existing checkpoints were not available online (EK-$100$ on TimeSformer, ViViT, and VideoMAE and SSv$2$ on ViViT), we finetune our own from existing K$400$ checkpoints on four GH$200$ GPUs~\cite{mcintoshsmith2024}.
Proportional attention is used in all models except VideoMAE, due to the model being a masked autoencoder derivative~\cite{bolya2022token}. 
For fair comparison, we use the same data loading/data augmentation techniques across all methods and each clip is only used for \emph{one} view, in contrast with the multiple views often used in the original papers.

\myparagraph{Merging Schedule}
In practice, when $r$ is mentioned in this section, we are referring to setting $r_i$ for each layer, according to one of the following schedules:
\textbf{Constant:} fixed $r$ per layer, denoted as ViViT.
\textbf{Decreasing:} $2r \rightarrow 0 $ per layer, denoted as ViViT$\downarrow$.
\textbf{Increasing:} $0 \rightarrow 2r$ per layer, denoted as ViViT$\uparrow$.
For the decreasing and increasing schedules, values are linearly interpolated between the two listed numbers.
For the divided space-time models, the $r$ value is a tuple of the $r$ value for each frame and the effective number of frames in each clip.

\myparagraph{Baselines}
We re-implement previous methods for token merging and dropout with our combination of models and datasets as baselines to compare with spatio-temporal token merging.
\textbf{Random dropout~\cite{liu2023patchdropout}:} Randomly drop tokens (instead of merging).
\textbf{Dropout~\cite{bolya2022token}:} Use attention metric to choose which tokens to drop (instead of merging).
Finally, we also introduce a naive baseline to determine how important the attention metric is when merging:
\textbf{Random merge:} Randomly merge tokens instead of using attention to choose.

\subsection{Dataset Details}
\label{sub:dataset_details}

We evaluate on three action recognition datasets: Kinetics-$400$ (K$400$)~\cite{kay2017kinetics}, Something-Something v$2$ (SSv$2$)~\cite{goyal2017something} and EPIC-KITCHENS-$100$ (EK-$100$)~\cite{damen2022rescaling}. K$400$ is a coarse-grained action recognition dataset collected from YouTube, focusing on a large range of general human actions, typically from a third-person perspective. Comparatively, SSv$2$ is more fine-grained and studies interactions between hands and objects. Most notably, the action classes are object agnostic, to encourage understanding of temporal cues. Lastly, EK-$100$ consists of unscripted egocentric footage gathered in participants' kitchens, which introduces increased visual noise in the form of motion blur and occlusions. Actions are annotated with a ``verb'' and ``noun'' class, making the dataset considerably more fine-grained.

\begin{table*}[htp]
\centering
\begin{tabular}{cccccccccc}
\hline
 &  &  &  &  & \multicolumn{3}{c}{EK-100} &  & Speedup \\ \cline{6-8}
\multirow{-2}{*}{Model} & \multirow{-2}{*}{$r$} & \multirow{-2}{*}{Reduction} & \multirow{-2}{*}{K400} & \multirow{-2}{*}{SSv2} & Action & Verb & Noun & \multirow{-2}{*}{FPS} & (X) \\ \hline
 & {\color[HTML]{9B9B9B} 0} & {\color[HTML]{9B9B9B} -} & {\color[HTML]{9B9B9B} 76.63} & {\color[HTML]{9B9B9B} 50.66} & {\color[HTML]{9B9B9B} 31.32} & {\color[HTML]{9B9B9B} 55.48} & {\color[HTML]{9B9B9B} 47.23} & {\color[HTML]{9B9B9B} 117.78} & {\color[HTML]{9B9B9B} 1.00} \\
 &  & random drop & 65.58 & 17.18 & 12.78 & 34.03 & 28.31 & 240.13 & 2.04 \\
 &  & drop & 68.30 & 22.97 & 16.09 & 38.24 & 33.02 & 240.13 & 2.04 \\
 &  & random merge & 38.41 & 5.46 & 2.82 & 21.47 & 8.57 & 234.58 & 1.99 \\
\multirow{-5}{*}{TimeSformer~\cite{bertasius2021space}} & \multirow{-4}{*}{18 $\times$ 8} & merge & \textbf{71.14} & \textbf{25.11} & \textbf{18.59} & \textbf{40.47} & \textbf{35.73} & 240.16 & 2.04 \\ \hline
 & {\color[HTML]{9B9B9B} 0} & {\color[HTML]{9B9B9B} -} & {\color[HTML]{9B9B9B} 70.50} & {\color[HTML]{9B9B9B} 61.39} & {\color[HTML]{9B9B9B} 35.02} & {\color[HTML]{9B9B9B} 61.09} & {\color[HTML]{9B9B9B} 46.72} & {\color[HTML]{9B9B9B} 99.79} & {\color[HTML]{9B9B9B} 1.00} \\
 &  & random drop & 63.80 & \textbf{24.50} & 14.27 & 37.38 & 27.57 & 218.40 & 2.19 \\
 &  & drop & 63.41 & 22.46 & \textbf{15.92} & \textbf{39.54} & \textbf{30.60} & 216.73 & 2.17 \\
 &  & random merge & 49.30 & 17.76 & 7.88 & 31.56 & 16.77 & 210.30 & 2.11 \\
\multirow{-5}{*}{Motionformer~\cite{patrick2021keeping}} & \multirow{-4}{*}{18 $\times$ 8} & merge & \textbf{65.05} & 24.10 & 15.60 & 38.20 & 30.15 & 218.11 & 2.19 \\ \hline
 & {\color[HTML]{9B9B9B} 0} & {\color[HTML]{9B9B9B} -} & {\color[HTML]{9B9B9B} 62.09} & {\color[HTML]{9B9B9B} 64.58} & {\color[HTML]{9B9B9B} 35.70} & {\color[HTML]{9B9B9B} 61.49} & {\color[HTML]{9B9B9B} 46.89} & {\color[HTML]{9B9B9B} 186.72} & {\color[HTML]{9B9B9B} 1.00} \\
 &  & random drop & 55.02 & 57.29 & 28.53 & 55.45 & 39.45 & 481.45 & 2.58 \\
 &  & drop & \textbf{56.65} & 60.33 & 31.02 & 57.70 & \textbf{42.43} & 483.04 & 2.59 \\
 &  & random merge & 20.64 & 22.89 & 5.44 & 26.86 & 10.07 & 471.74 & 2.53 \\
\multirow{-5}{*}{VideoMAE~\cite{tong2022videomae}} & \multirow{-4}{*}{150} & merge & 56.10 & \textbf{61.10} & \textbf{31.27} & \textbf{58.00} & 42.39 & 476.28 & 2.55 \\ \hline
 & {\color[HTML]{9B9B9B} 0} & {\color[HTML]{9B9B9B} -} & {\color[HTML]{9B9B9B} 63.43} & {\color[HTML]{9B9B9B} 50.63} & {\color[HTML]{9B9B9B} 35.82} & {\color[HTML]{9B9B9B} 58.19} & {\color[HTML]{9B9B9B} 51.59} & {\color[HTML]{9B9B9B} 106.00} & {\color[HTML]{9B9B9B} 1.00} \\
 &  & random drop & 59.95 & 46.71 & 30.36 & 54.24 & 45.51 & 262.04 & 2.47 \\
 &  & drop & 58.00 & 45.36 & 30.12 & 53.20 & 46.90 & 262.34 & 2.47 \\
 &  & random merge & 28.88 & 19.15 & 5.78 & 28.88 & 14.82 & 259.92 & 2.45 \\
\multirow{-5}{*}{ViViT~\cite{arnab2021vivit}} & \multirow{-4}{*}{300} & merge & \textbf{63.08} & \textbf{50.15} & \textbf{35.11} & \textbf{57.24} & \textbf{51.33} & 260.72 & 2.46 \\ \hline
\end{tabular}
\caption{Performance of token merging with a constant schedule when compared to alternative methods of reducing token sequence length. Bold indicates the reduction methods that achieve highest accuracy on a given dataset. Grey rows correspond to the upper bound accuracy of the original model.}
\label{tab:results}
\end{table*}

\subsection{Quantitative Results}
\label{sub:quantitative_results}

\myparagraph{Scaling with $r$}
We first investigate the trade-off between accuracy and throughput from token merging by varying $r$. 
The results can be seen in~\cref{fig:scaling_curves}, which compares the Kinetics-$400$ accuracy and speedup gained for these schedules on ViViT and VideoMAE.
Note that we plot $r$ as a percentage of tokens dropped for easy comparison between the two models.

In~\cref{fig:scaling_curves_a}, looking at the constant schedule, we see that both ViViT and VideoMAE are resilient to merging up to $10\%$ of their original tokens \textit{per layer}, as after this point the accuracy (especially for VideoMAE) begins to decrease significantly. 
Nevertheless, this still represents dropping a total of $60\%$ tokens throughout the entire network.
At the other extreme, when $r$ is $20\%$ of the original tokens, VideoMAE sinks to almost random accuracy, while ViViT retains much more of its original performance.
The increasing schedule displays a significantly slower drop in accuracy, while the decreasing schedule drops significantly faster becaused merging is loaded towards the front of the model.
However, when using the decreasing schedule, there is still a period before $r$ hits $10\%$ of the tokens where the accuracy matches that of the more conservative schedules.

We compare model accuracy to the corresponding speedup gained via merging in~\cref{fig:scaling_curves_b}, where we see an elbow for each curve, indicating the points at which merging stops being economical in terms of throughput.
The increasing schedules with loosely dashed lines are gathered together, showing little gain in speedup, while the tightly dashed lines for the decreasing schedule show quicker gains in speedup.
The constant schedule is a middle-ground between the two, reaching the same endpoint as the decreasing schedule at a slower rate.
Finally, in terms of picking optimal $r$ and schedule, the point at each elbow indicates that for the constant and decreasing schedules, merging $10\%$ of the original tokens can produce speedups of roughly $2.5$X and $4$X respectively.

From inspection of these results, we select $r$ for all other experiments such that $10\%$ of tokens are dropped per layer, equating to $60\%$ of all tokens across all layers in the network being merged.
This value gives a good trade-off between retaining accuracy but increasing throughput of the methods.

\begin{figure*}[!t]
  \centering
  \includegraphics[width=0.97\linewidth]{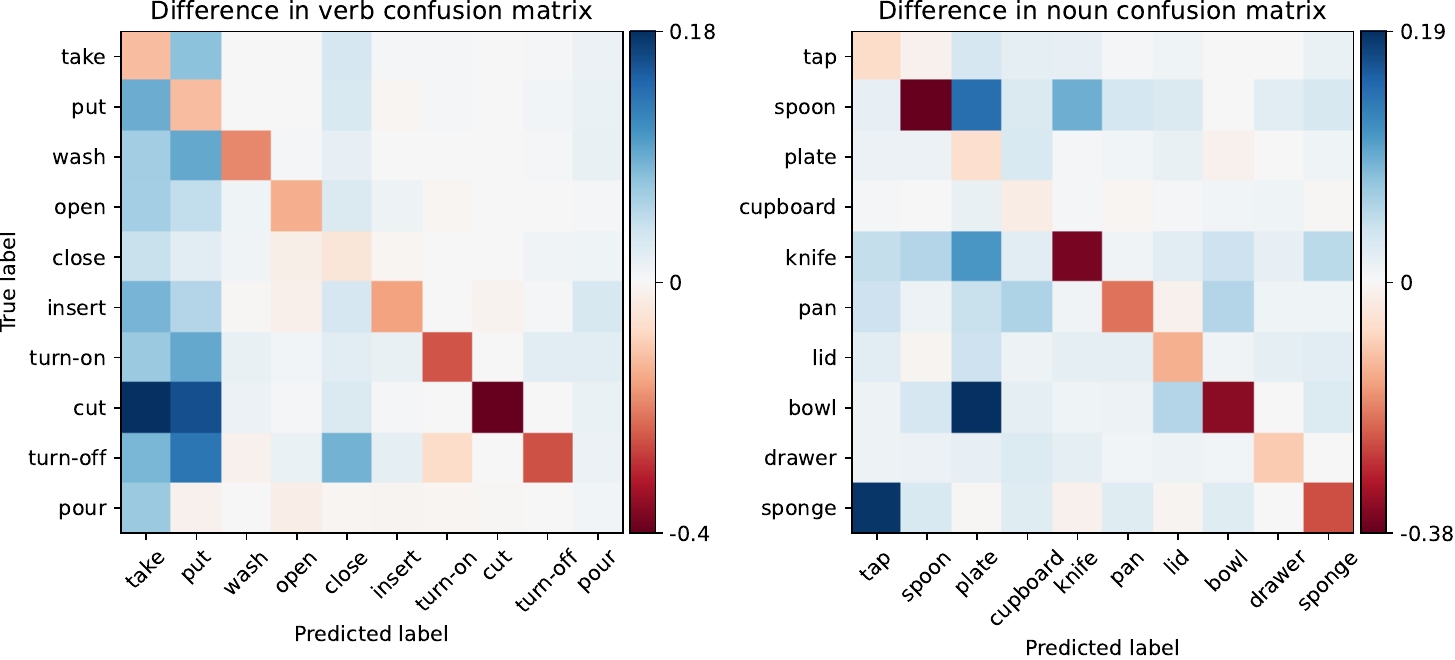}
  \caption{Impact on confusion matrices from Token Merging a VideoMAE model. The first $10$ verb and noun classes are displayed left and right respectively from VideoMAE on EK-$100$. Red indicates less predictions and blue indicates more predictions.}
  \label{fig:confusion_difference}
\end{figure*}

\begin{figure*}[htp]
  \centering
  \begin{subfigure}{1.0\linewidth}
    \centering
    \includegraphics[width=1.0\linewidth]{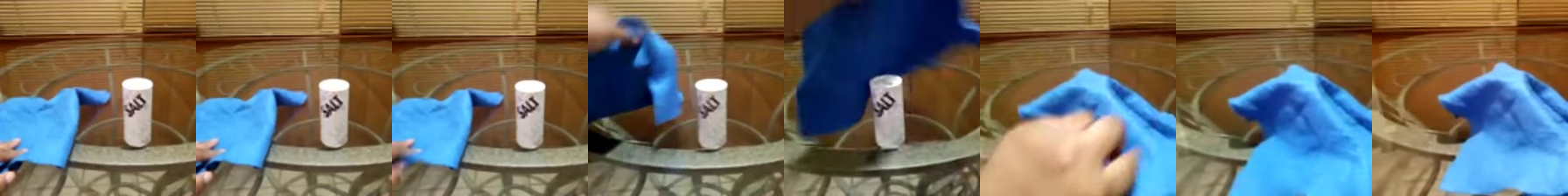}
    \caption{Original clip.}
    \label{fig:ssv2_example_a}
  \end{subfigure}
  \hfill
  \begin{subfigure}{1.0\linewidth}
    \centering
    \includegraphics[width=1.0\linewidth]{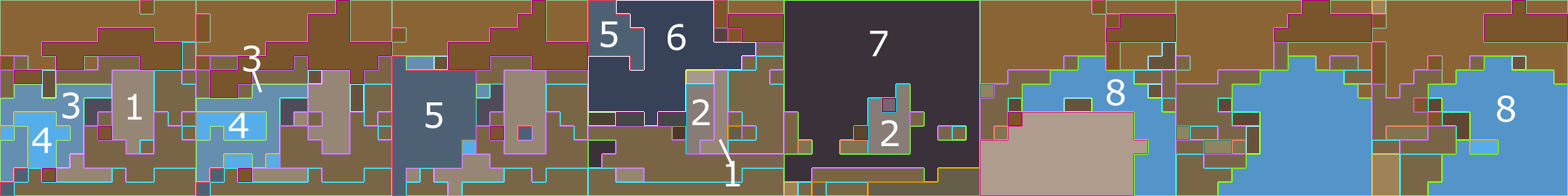}
    \caption{Merged clip.}
    \label{fig:ssv2_example_b}
  \end{subfigure}
  \caption{Visualisation of the final merged tokens for an SSv$2$ clip of ``covering salt shaker with a towel'', produced with VideoMAE. Tokens $1$ and $2$ capture the white salt shaker. The model struggles more with the blue towel, with it splitting into tokens $3$ -- $8$.}
  \label{fig:ssv2_example}
\end{figure*}

\begin{figure*}[htp]
  \centering
  \begin{subfigure}{1.0\linewidth}
    \centering
    \includegraphics[width=1.0\linewidth]{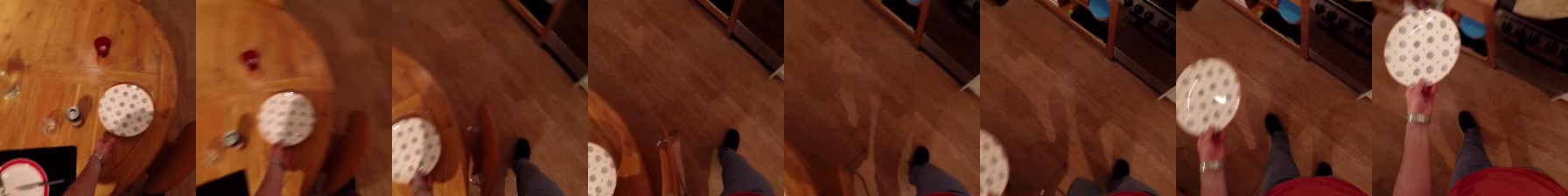}
    \caption{Original clip.}
    \label{fig:epickitchens_example_a}
  \end{subfigure}
  \hfill
  \begin{subfigure}{1.0\linewidth}
    \centering
    \includegraphics[width=1.0\linewidth]{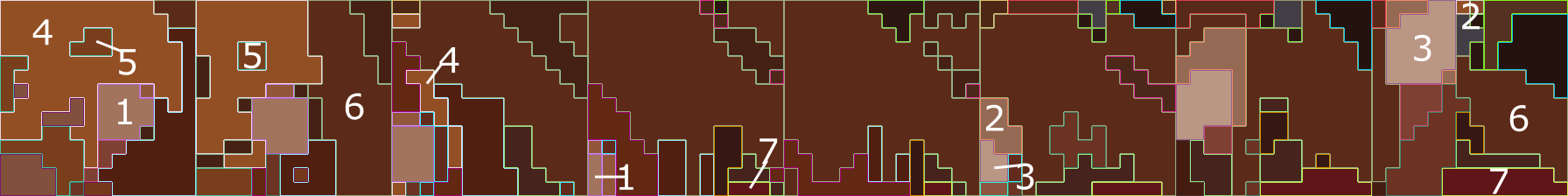}
    \caption{Merged clip.}
    \label{fig:epickitchens_example_b}
  \end{subfigure}
  \caption{Visualisation of the final merged tokens for an EK-$100$ clip of ``take plate'', produced with VideoMAE. The first $4$ frames, merge the plate into token $1$, even with motion blur whereas in the last $3$ frames it splits into tokens $2$ and $3$.}
  \label{fig:epickitchens_example}
\end{figure*}

\begin{figure*}[htp]
  \centering
  \begin{subfigure}{1.0\linewidth}
    \centering
    \includegraphics[width=1.0\linewidth]{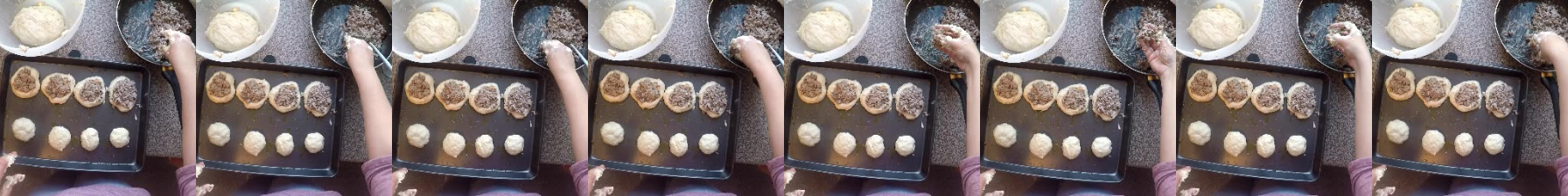}
    \caption{Original clip.}
    \label{fig:epickitchens_duplicate_layer_a}
  \end{subfigure}
  \hfill
  \begin{subfigure}{1.0\linewidth}
    \centering
    \includegraphics[width=1.0\linewidth]{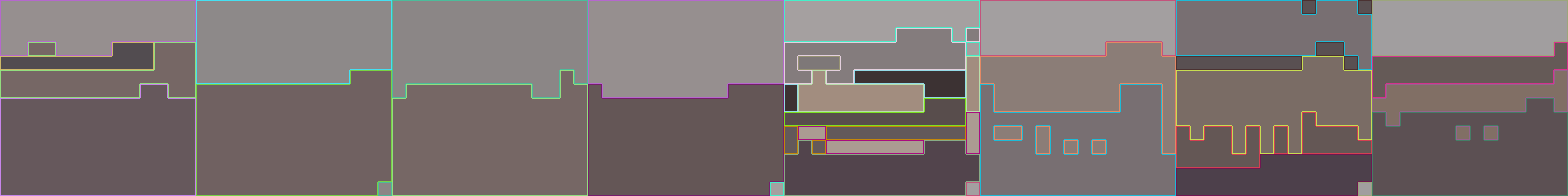}
    \caption{Merged clip when duplicating layer $1$.}
    \label{fig:epickitchens_duplicate_layer_b}
  \end{subfigure}
  \hfill
  \begin{subfigure}{1.0\linewidth}
    \centering
    \includegraphics[width=1.0\linewidth]{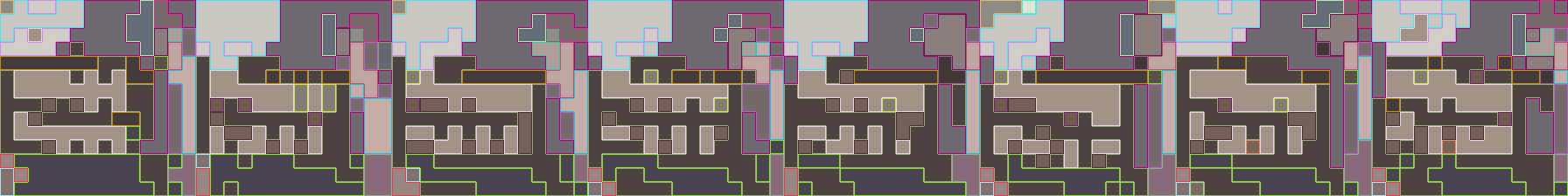}
    \caption{Merged clip when duplicating layer $12$.}
    \label{fig:epickitchens_duplicate_layer_c}
  \end{subfigure}
  \caption{Visualisations of the difference in merging decisions made in layer $1$ versus layer $12$, produced with VideoMAE. We find that the final layer merges the tokens in a more object focused manner vs. the first layer.}
  \label{fig:epickitchens_duplicate_layer}
\end{figure*}

\begin{figure*}[htp]
  \centering
  \begin{subfigure}{1.0\linewidth}
    \centering
    \includegraphics[width=1.0\linewidth]{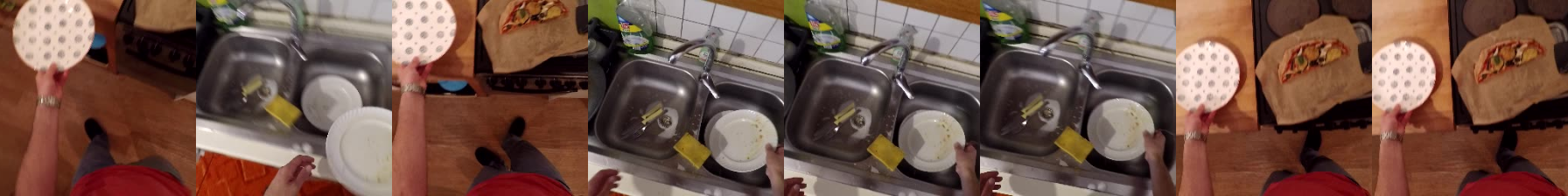}
    \caption{Spliced clip.}
    \label{fig:epickitchens_splicing_example_a}
  \end{subfigure}
  \hfill
  \begin{subfigure}{1.0\linewidth}
    \centering
    \includegraphics[width=1.0\linewidth]{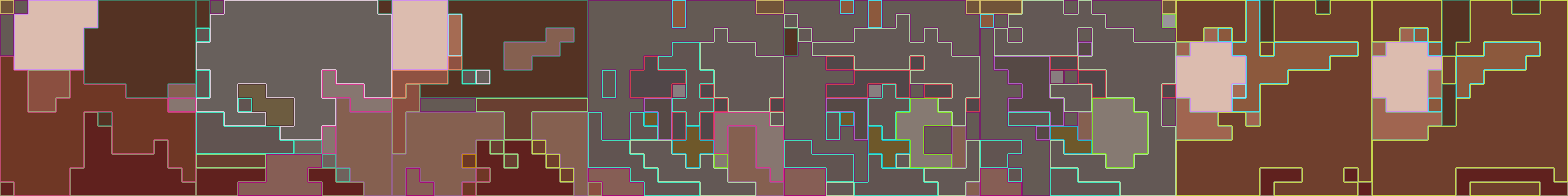}
    \caption{Merged clip.}
    \label{fig:epickitchens_splicing_example_b}
  \end{subfigure}
  \caption{Merging outcome for a clip that has had half its frames from the most ``similar'' clip in the same noun class spliced in, produced with VideoMAE. The plates are merged consistently together within each clip but are not merged across the clips.}
  \label{fig:epickitchens_splicing_example}
\end{figure*}

\myparagraph{Comparison with token reduction methods}
In~\cref{tab:results}, we show results across all models \emph{without} any reduction and \emph{with} reduction at a specified $r$ value, when using a constant schedule. 
We compare upper bound accuracy (in grey) with four reduction methods which don't require re-training, including our spatio-temporal merging.

Firstly, we investigate the importance of the merging aspect by comparing between reduction methods. 
For all methods but Motionformer, token merging remains able to produce the highest accuracies when reducing token sequences. 
On all datasets tested, ViViT gains at least $4\%$  over dropout that uses the same attention metric. 
For VideoMAE, the gains are less significant, yet consistently outperform other methods on SSv$2$ and EK-$100$.
An interesting result is that randomly merging tokens is significantly destructive to performance, being noticeably worse than random dropout. 
In particular, on EK-$100$, the verb accuracy tends to halve, while the noun accuracy roughly quarters.
The poor performance can be explained by the fact that potentially dissimilar tokens are having their features averaged together, creating a more negative effect than averaging the features of tokens that the model already determines to be similar---in this case dropping the tokens altogether is preferable.
We note that there are not significant differences between the throughputs of these variations.

Secondly, we directly compare the upper bound model performance to our merged implementations.
For the divided space-time models, accuracy on K$400$ drops by roughly $5\%$, while performance on SSv$2$ and EK-$100$ drops by much larger margins. 
In particular, the accuracies on SSv$2$ are roughly half what the vanilla model can achieve even with a speedup of just over $2$X.
Both EK-$100$ and SSv$2$ are more fine-grained and reliant on temporal reasoning, \textit{token merging is impairing the models' ability to fuse information across frames}, while high accuracy can be attained on K$400$ with spatial reasoning. 

We find that spatio-temporal models are much more resilient to merging. 
VideoMAE sees losses of around $4\%$ across all datasets, though it retains strong accuracy on SSv$2$ and EK-$100$, either outperforming or being comparable to $r=0$ TimeSformer and Motionformer. 
ViViT with token merging \textit{performs extremely well on all datasets}, never dropping by more than $1\%$.
Both models see speedups of around $2.5$X, with ViViT reaching comparable performance, \textit{increasing the inference throughput for free}.
While ViViT and VideoMAE have a similar structure, there are a few differences that could account for the disparity: ViViT uses twice as many input frames whereas VideoMAE does not use a class token and is pretrained on masked input.

\myparagraph{Comparison of class confusion}
We investigate the errors introduced from merging VideoMAE and produce confusion matrices for the most frequent $10$ verb and noun classes in EK-$100$. 
\Cref{fig:confusion_difference} shows the difference in performance between a model \textit{with} and a model \textit{without} token merging.
In~\cref{fig:confusion_difference} (left) we can see that when a high $r$ value is used, the merged model becomes more likely to predict the highly common ``take'' and ``put'' classes, exacerbating the errors of the original model.

The nouns in~\cref{fig:confusion_difference} (right) are more resilient to this collapse into the most common classes.
However, we can see that certain related nouns are increasingly being misclassified, particularly those that relate to smaller objects.
For example, ``spoon'' and ``knife'' are more frequently misclassified as the wrong piece of cutlery, or the action-related ``plate''. 
Similarly, ``bowl'' is mistaken for ``plate'' and ``sponge'' is mistaken for ``tap''.
Larger objects like ``drawer'', ``tap'', and ``cupboard''---which occupy portions of the background---are broadly unaffected by the merging, suggesting that the features of smaller objects are more likely to be lost by merging.

\subsection{Qualitative Results}
\label{sub:qualitative_results}

In this section, we explore the token merging from a visual standpoint, answering the following questions: i) How are tokens merged in video?; ii) How do different layers merge tokens?; and iii) Are tokens being merged semantically or visually?
For the visualisations, we extended the implementation in~\cite{bolya2022token} for video---tokens with the same background and edge colour are the same. 
We number the first and last appearance in each visualisation where specific tokens are referred to in text.

\noindent \textbf{How are tokens merged in video?}
The action ``covering salt shaker with a towel'' from SSv$2$ is depicted in~\cref{fig:ssv2_example}. 
While it is visible, the white salt shaker is captured well by token $1$ and $2$, likely split due to the shadow of the towel covering the salt shaker. 
However, the model struggles with the blue towel, leaving tokens $3$--$8$ unmerged. 
Tokens $4$ and $5$ are significantly darker as the towel's underside is not hit by a light source, increasing the visual differences between the tokens, suggesting that tokens are not merged semantically and visual differences will prevent merging across frames.

Next, we show an example from EK-$100$ of ``take plate'' in~\cref{fig:epickitchens_example}. 
In the first four frames, the plate is merged to token $1$, even with the rapid motion blur. 
However, when it's back in view in the last three frames it splits into tokens $2$ and $3$. 
Additionally, the table and a glass are clearly grouped separately into tokens $4$ and $5$ respectively. 
The background is surprisingly well distinguished from the foreground in token $6$, even with the unusual head pose, likely because the floor has one common texture. 
Interestingly, the t-shirt of the participant is visibly grouped into token $7$.
These examples showcase how tokens are merged effectively spatio-temporally without any learned parameters/fine-tuning.

\noindent \textbf{How do different layers merge tokens?}
Because the merging method happens iteratively as the token sequence passes through transformer layers, the outcome is cumulative and biased towards earlier layers. 
To determine how different transformer layers merge tokens, we conduct experiments where layer $i$ of the model is duplicated $12$ times (stripping out everything but the merging modules) and tokens are only merged by instances of layer $i$. 
\Cref{fig:epickitchens_duplicate_layer} visualises an example from EK-$100$. 
In~\cref{fig:epickitchens_duplicate_layer_b} we see that duplicating the first layer produces merged results that are drastically different compared to earlier examples. 
We speculate that this layer is merging based on rudimentary shape or basic foreground/background colours.
In~\cref{fig:epickitchens_duplicate_layer_c}, we can see that duplicating the final layer produces similar output to a visual segementation of the clip. 
The food, tray, bowl, and background are clearly separated consistently across frames.
Due to the cumulative nature of the layer based merging, there is currently no simple method to balance the impact that different layers have.
However, \textit{biasing merging towards the later layers significantly decreases the gains in throughput} of the model. 

\noindent \textbf{Are tokens being merged semantically or visually?}
Finally, we test whether the models are merged semantically, as much of what we have demonstrated is possible with visual similarity. 
We do this by taking a video and inserting clips from a different video from the same class, i.e. a clip with the same semantic meaning and objects and check how the tokens are merged spatially and temporally.
We find videos that the model believes to be semantically similar by comparing the noun logits of all clips in the EK-$100$ test set using the Kullback–Leibler (KL) divergence between all possible pairs, and select the nearest video. 
\Cref{fig:epickitchens_splicing_example} shows an example of the most ``similar'' corresponding clip in the same noun class, randomly splicing frames into the source video.
From the example, we can see that the model merges the original clip's plate into a single token, which notably does not overlap with any of the second clip's corresponding plate tokens. 
Even though the objects are in the same class, the difference in lighting and texture is enough to stop their tokens from merging, suggesting that \textit{the merging process is almost entirely visual}.
We show more examples within the supplementary material, but only a tiny portion of examples suggest that tokens are being merged with any semantic reasoning.
We also find that the background is only merged between the original and spliced frames if the kitchen is visually similar enough or the light levels are similar, indicating that the merging is not explicitly considering foreground and background.

\section{Conclusion}
\label{sec:conclusion}

The efficiency of transformer models remains an ongoing cornerstone of research in deep learning. 
In this work, we implemented video token merging for two spatio-temporal vision transformers and two divided space-time transformers to see how effectively the method can be dropped into existing models without re-training. 
As well as this, we created a testbed to compare token merging across video transformer models.

Using this testbed, we collate various quantitative and qualitative findings, with the aim of exploring how and why tokens are merged in the temporal dimension, as well as the effect that this has on action recognition with fine-grained video datasets.
We have determined the margin by which attention driven merging beats dropout alternatives.
Furthermore, we have demonstrated that this method of merging has little consideration for the semantics in the inputs that it merges and that the merging decisions made by different layers can vary wildly.
The token merging methodology remains a convenient drop-in for vision transformers to increase throughput with a small drop in performance based on the ratio of dropped tokens.
We hope that with these results pave the way for token merging to be directly incorporated into future video transformer models.
\section*{Acknowledgements}

Research supported by EPSRC Doctoral Training Partnerships (DTP).
The authors would like to thank Siddhant Bansal, Prajwal Gatti and Toby Perrett for their comments on the paper.
The authors acknowledge the use of resources provided by the Isambard-AI National AI Research Resource (AIRR). Isambard-AI is operated by the University of Bristol and is funded by the UK Government’s Department for Science, Innovation and Technology (DSIT) via UK Research and Innovation; and the Science and Technology Facilities Council [ST/AIRR/I-A-I/1023].

{
    \small
    \bibliographystyle{ieeenat_fullname}
    \bibliography{main}
}

\clearpage
\setcounter{page}{1}
\maketitlesupplementary

\noindent In the supplementary material we provide extra results for the speedup of different methods in \cref{sub:throughput_curve}, comparisons to other token reduction strategies for increasing/decreasing schedules in \cref{sub:inc_dec_schedule}, and confusion matrices for ViViT in \cref{sub:vivit_confusion}. We also provide information on training hyperparameters (for finetuning where it was required) in \cref{sec:training_hyperparameters} and many more qualitative figures in \cref{sec:qualitative_examples} including extra merging examples, visualisations of layer decisions, and tests of semantic merging.

\section{Additional Quantitative Results}
\label{sec:additional_results}

In this section, we introduce extra results that we were not able to include in the main paper. We provide further scaling curves, results tables and confusion matrices on Kinetics-$400$ (K$400$)~\cite{kay2017kinetics}, Something-Something v$2$ (SSv$2$)~\cite{goyal2017something} and EPIC-KITCHENS-$100$ (EK-$100$)~\cite{damen2022rescaling}.

\subsection{Throughput Curve}
\label{sub:throughput_curve}

\begin{figure}[htp]
  \centering
  \includegraphics[width=\linewidth]{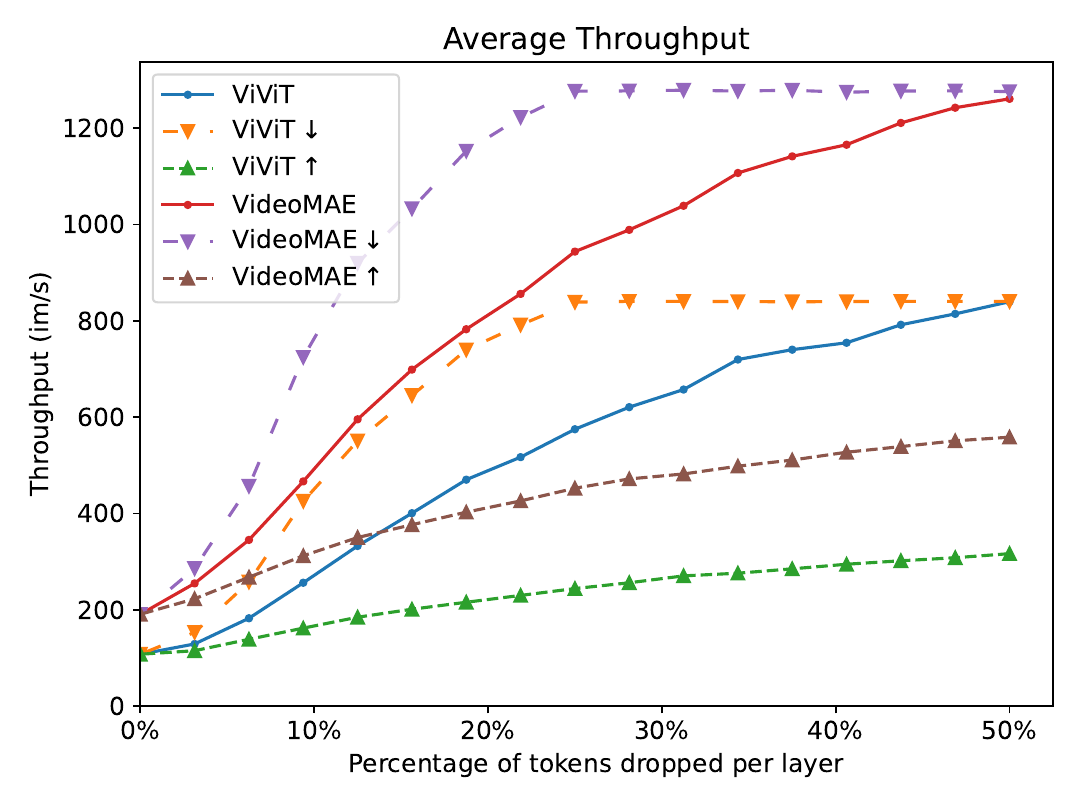}
  \caption{Curve corresponding to image throughput with ViViT and VideoMAE on K$400$ when increasing $r$ (the number of tokens merged) up to its limit. The $x$-axis is the percentage (relative to the original total) of tokens dropped \textit{per layer}.}
  \label{fig:throughput_curve}
\end{figure}

With~\cref{fig:throughput_curve}, we plot the throughput (in terms of images per second) of ViViT and VideoMAE on K$400$ for the constant, decreasing and increasing schedules. 
The increasing schedules introduce significantly slower speedups, showing that even with maximum merging per layer, it's only possible to introduce a speedup of roughly $2.5$X.
At the same proportion of merging, the decreasing and constant schedules meet at the same endpoints, achieving roughly $7$X for VideoMAE and $8$X for ViViT.

\subsection{Analysis of Increasing/Decreasing Schedules}
\label{sub:inc_dec_schedule}

\begin{table*}[htp]
\centering
\begin{tabular}{cccccccccc}
\hline
 &  &  &  &  & \multicolumn{3}{c}{EK-100} &  & Speedup \\ \cline{6-8}
\multirow{-2}{*}{Model} & \multirow{-2}{*}{$r$} & \multirow{-2}{*}{Reduction} & \multirow{-2}{*}{K400} & \multirow{-2}{*}{SSv2} & Action & Verb & Noun & \multirow{-2}{*}{FPS} & (X) \\ \hline
 & {\color[HTML]{9B9B9B} 0} & {\color[HTML]{9B9B9B} -} & {\color[HTML]{9B9B9B} 76.63} & {\color[HTML]{9B9B9B} 50.66} & {\color[HTML]{9B9B9B} 31.32} & {\color[HTML]{9B9B9B} 55.48} & {\color[HTML]{9B9B9B} 47.23} & {\color[HTML]{9B9B9B} 117.78} & {\color[HTML]{9B9B9B} 1.00} \\
 &  & random drop & 28.38 & 9.80 & 1.38 & 18.68 & 6.12 & 361.01 & 3.07 \\
 &  & drop & \textbf{30.41} & \textbf{11.23} & \textbf{1.68} & 19.45 & \textbf{6.36} & 359.59 & 3.05 \\
 &  & random merge & 3.32 & 1.72 & 0.85 & 17.07 & 2.76 & 354.71 & 3.01 \\
\multirow{-5}{*}{TimeSformer~\cite{bertasius2021space}} & \multirow{-4}{*}{18 $\times$ 8} & merge & 25.26 & 9.22 & 1.39 & \textbf{19.90} & 5.80 & 360.33 & 3.06 \\ \hline
 & {\color[HTML]{9B9B9B} 0} & {\color[HTML]{9B9B9B} -} & {\color[HTML]{9B9B9B} 70.50} & {\color[HTML]{9B9B9B} 61.39} & {\color[HTML]{9B9B9B} 35.02} & {\color[HTML]{9B9B9B} 61.09} & {\color[HTML]{9B9B9B} 46.72} & {\color[HTML]{9B9B9B} 99.79} & {\color[HTML]{9B9B9B} 1.00} \\
 &  & random drop & 46.27 & 20.68 & 8.31 & 31.58 & 17.10 & 331.00 & 3.32 \\
 &  & drop & 48.53 & 21.66 & \textbf{10.14} & \textbf{33.70} & \textbf{19.47} & 329.94 & 3.31 \\
 &  & random merge & 17.12 & 6.24 & 1.58 & 22.23 & 5.35 & 328.19 & 3.29 \\
\multirow{-5}{*}{Motionformer~\cite{patrick2021keeping}} & \multirow{-4}{*}{18 $\times$ 8} & merge & \textbf{50.64} & \textbf{21.73} & 8.91 & 33.18 & 18.05 & 334.57 & 3.35 \\ \hline
 & {\color[HTML]{9B9B9B} 0} & {\color[HTML]{9B9B9B} -} & {\color[HTML]{9B9B9B} 62.09} & {\color[HTML]{9B9B9B} 64.58} & {\color[HTML]{9B9B9B} 35.70} & {\color[HTML]{9B9B9B} 61.49} & {\color[HTML]{9B9B9B} 46.89} & {\color[HTML]{9B9B9B} 186.72} & {\color[HTML]{9B9B9B} 1.00} \\
 &  & random drop & 20.48 & 24.17 & 10.24 & 34.93 & 17.56 & 748.05 & 4.01 \\
 &  & drop & \textbf{23.08} & 31.30 & \textbf{11.99} & \textbf{37.91} & \textbf{19.86} & 747.32 & 4.00 \\
 &  & random merge & 1.04 & 2.99 & 0.57 & 14.60 & 2.11 & 735.87 & 3.94 \\
\multirow{-5}{*}{VideoMAE~\cite{tong2022videomae}} & \multirow{-4}{*}{150} & merge & 20.82 & \textbf{33.34} & 10.88 & 36.31 & 18.06 & 742.91 & 3.98 \\ \hline
 & {\color[HTML]{9B9B9B} 0} & {\color[HTML]{9B9B9B} -} & {\color[HTML]{9B9B9B} 63.43} & {\color[HTML]{9B9B9B} 50.63} & {\color[HTML]{9B9B9B} 35.82} & {\color[HTML]{9B9B9B} 58.19} & {\color[HTML]{9B9B9B} 51.59} & {\color[HTML]{9B9B9B} 106.00} & {\color[HTML]{9B9B9B} 1.00} \\
 &  & random drop & 42.71 & 29.58 & 13.16 & 37.33 & 25.11 & 436.39 & 4.12 \\
 &  & drop & 43.94 & 32.30 & 14.92 & 38.93 & 28.69 & 433.73 & 4.09 \\
 &  & random merge & 2.67 & 2.03 & 0.73 & 17.64 & 4.14 & 432.76 & 4.08 \\
\multirow{-5}{*}{ViViT~\cite{arnab2021vivit}} & \multirow{-4}{*}{300} & merge & \textbf{57.01} & \textbf{43.80} & \textbf{23.78} & \textbf{48.79} & \textbf{37.55} & 439.26 & 4.14 \\ \hline
\end{tabular}
\caption{Performance of token merging with a decreasing schedule when compared to alternative methods of reducing token sequence length. Bold indicates the reduction methods that achieve highest accuracy on a given dataset. Grey rows correspond to the upper bound accuracy of the original model.}
\label{tab:decreasing_results_table}
\end{table*}

In \cref{tab:decreasing_results_table}, we have a table of results comparing reduction strategies for the decreasing schedule, using the same $r$ value as in the main paper.
Comparing across reduction strategies, we see that biasing merging towards the earlier layers is resulting in attention based dropout performing better than token merging for TimeSformer, Motionformer and VideoMAE.
Random merging is essentially unuseable in this scenario, with accuracies approaching random performance in most metrics.
Next, we directly compare the upper bound accuracy of the original models to those implementing a decreasing merging schedule.
All models display large drops in accuracy, with VideoMAE in particular dropping by almost $50\%$ on K$400$, $30\%$ on SSv$2$ and $45\%$ on EK-$100$.
We've demonstrated that early transformer layers make merging decisions that do not correspond to visual segmentations (see~\cref{sec:qualitative_examples} for more examples), suggesting that biasing merging towards earlier layers may introduce merging that quickly obfuscates visual features.
ViViT retains more accuracy, especially on EK-$100$, where it drops by roughly $5\%$ across the different label types.
Looking to the speedup gained, we see that the throughput has been increased significantly to a $4$X speedup, from the $2.5$X speedup that the constant schedule demonstrates in the main paper.
From this table, we can determine that a lower $r$ value is required for the decreasing schedule, otherwise merging begins to be especially detrimental for models other than ViViT.

\begin{table*}[htp]
\centering
\begin{tabular}{cccccccccc}
\hline
 &  &  &  &  & \multicolumn{3}{c}{EK-100} &  & Speedup \\ \cline{6-8}
\multirow{-2}{*}{Model} & \multirow{-2}{*}{$r$} & \multirow{-2}{*}{Reduction} & \multirow{-2}{*}{K400} & \multirow{-2}{*}{SSv2} & Action & Verb & Noun & \multirow{-2}{*}{FPS} & (X) \\ \hline
 & {\color[HTML]{9B9B9B} 0} & {\color[HTML]{9B9B9B} -} & {\color[HTML]{9B9B9B} 76.63} & {\color[HTML]{9B9B9B} 50.66} & {\color[HTML]{9B9B9B} 31.32} & {\color[HTML]{9B9B9B} 55.48} & {\color[HTML]{9B9B9B} 47.23} & {\color[HTML]{9B9B9B} 117.78} & {\color[HTML]{9B9B9B} 1.00} \\
 &  & random drop & 72.36 & 22.03 & 19.36 & 40.29 & 37.73 & 163.89 & 1.39 \\
 &  & drop & 72.72 & 27.54 & 21.98 & 43.78 & 39.88 & 166.17 & 1.41 \\
 &  & random merge & 65.96 & 16.27 & 14.63 & 37.46 & 28.75 & 163.91 & 1.39 \\
\multirow{-5}{*}{TimeSformer~\cite{bertasius2021space}} & \multirow{-4}{*}{18 $\times$ 8} & merge & \textbf{74.24} & \textbf{28.91} & \textbf{23.28} & \textbf{45.23} & \textbf{41.56} & 163.97 & 1.39 \\ \hline
 & {\color[HTML]{9B9B9B} 0} & {\color[HTML]{9B9B9B} -} & {\color[HTML]{9B9B9B} 70.50} & {\color[HTML]{9B9B9B} 61.39} & {\color[HTML]{9B9B9B} 35.02} & {\color[HTML]{9B9B9B} 61.09} & {\color[HTML]{9B9B9B} 46.72} & {\color[HTML]{9B9B9B} 99.79} & {\color[HTML]{9B9B9B} 1.00} \\
 &  & random drop & 67.79 & 31.07 & 20.06 & 43.29 & 35.62 & 142.05 & 1.42 \\
 &  & drop & 67.56 & 31.74 & 22.60 & 46.28 & 38.36 & 143.44 & 1.44 \\
 &  & random merge & 64.86 & 30.06 & 18.04 & 41.54 & 32.37 & 142.56 & 1.43 \\
\multirow{-5}{*}{Motionformer~\cite{patrick2021keeping}} & \multirow{-4}{*}{18 $\times$ 8} & merge & \textbf{68.00} & \textbf{32.98} & \textbf{22.91} & \textbf{46.39} & \textbf{38.73} & 141.77 & 1.42 \\ \hline
 & {\color[HTML]{9B9B9B} 0} & {\color[HTML]{9B9B9B} -} & {\color[HTML]{9B9B9B} 62.09} & {\color[HTML]{9B9B9B} 64.58} & {\color[HTML]{9B9B9B} 35.70} & {\color[HTML]{9B9B9B} 61.49} & {\color[HTML]{9B9B9B} 46.89} & {\color[HTML]{9B9B9B} 186.72} & {\color[HTML]{9B9B9B} 1.00} \\
 &  & random drop & 60.09 & 62.53 & 33.15 & 59.70 & 44.41 & 312.60 & 1.67 \\
 &  & drop & 60.44 & 63.59 & \textbf{34.35} & \textbf{60.56} & \textbf{45.85} & 319.38 & 1.71 \\
 &  & random merge & 48.32 & 55.00 & 22.91 & 50.56 & 33.15 & 316.12 & 1.69 \\
\multirow{-5}{*}{VideoMAE~\cite{tong2022videomae}} & \multirow{-4}{*}{150} & merge & \textbf{60.43} & \textbf{63.66} & 34.34 & 60.28 & 45.24 & 311.66 & 1.67 \\ \hline
 & {\color[HTML]{9B9B9B} 0} & {\color[HTML]{9B9B9B} -} & {\color[HTML]{9B9B9B} 63.43} & {\color[HTML]{9B9B9B} 50.63} & {\color[HTML]{9B9B9B} 35.82} & {\color[HTML]{9B9B9B} 58.19} & {\color[HTML]{9B9B9B} 51.59} & {\color[HTML]{9B9B9B} 106.00} & {\color[HTML]{9B9B9B} 1.00} \\
 &  & random drop & 62.53 & 49.59 & 34.15 & 57.04 & 49.63 & 165.18 & 1.56 \\
 &  & drop & 60.32 & 48.11 & 32.87 & 55.39 & 49.61 & 165.59 & 1.56 \\
 &  & random merge & 51.94 & 37.41 & 21.25 & 6.33 & 36.07 & 164.64 & 1.55 \\
\multirow{-5}{*}{ViViT~\cite{arnab2021vivit}} & \multirow{-4}{*}{300} & merge & \textbf{63.18} & \textbf{50.52} & \textbf{35.86} & \textbf{57.99} & \textbf{51.69} & 164.13 & 1.55 \\ \hline
\end{tabular}
\caption{Performance of token merging with an increasing schedule when compared to alternative methods of reducing token sequence length. Bold indicates the reduction methods that achieve highest accuracy on a given dataset. Grey rows correspond to the upper bound accuracy of the original model.}
\label{tab:increasing_results_table}
\end{table*}

\begin{table*}[htp]
\centering
\begin{tabular}{ccccccccc}
\hline
 &  &  &  &  &  & \multicolumn{3}{c}{EK-100} \\ \cline{7-9} 
\multirow{-2}{*}{Model} & \multirow{-2}{*}{$r$} & \multirow{-2}{*}{$t$} & \multirow{-2}{*}{Reduction} & \multirow{-2}{*}{K400} & \multirow{-2}{*}{SSv2} & Action & Verb & Noun \\ \hline
 & {\color[HTML]{9B9B9B} 0} & - & {\color[HTML]{9B9B9B} -} & {\color[HTML]{9B9B9B} 62.09} & {\color[HTML]{9B9B9B} 64.58} & {\color[HTML]{9B9B9B} 35.70} & {\color[HTML]{9B9B9B} 61.49} & {\color[HTML]{9B9B9B} 46.89} \\
 &  & - & merge & 56.10 & \textbf{61.10} & 31.27 & \textbf{58.00} & 42.39 \\
\multirow{-3}{*}{VideoMAE~\cite{tong2022videomae}} & \multirow{-2}{*}{150} & 0.8 & hybrid & \textbf{56.53} & 61.04 & \textbf{31.62} & 57.90 & \textbf{42.88} \\ \hline
 & {\color[HTML]{9B9B9B} 0} & - & {\color[HTML]{9B9B9B} -} & {\color[HTML]{9B9B9B} 63.43} & {\color[HTML]{9B9B9B} 50.63} & {\color[HTML]{9B9B9B} 35.82} & {\color[HTML]{9B9B9B} 58.19} & {\color[HTML]{9B9B9B} 51.59} \\
 &  & - & merge & 63.08 & 50.15 & 35.11 & 57.24 & \textbf{51.33} \\
\multirow{-3}{*}{ViViT~\cite{arnab2021vivit}} & \multirow{-2}{*}{300} & 0.4 & hybrid & \textbf{63.09} & \textbf{50.15} & \textbf{35.21} & \textbf{57.48} & 51.30 \\ \hline
\end{tabular}
\caption{Performance of hybrid token merging with a constant schedule when compared to vanilla token merging. Bold indicates the reduction methods that achieve highest accuracy on a given dataset. Grey rows correspond to the upper bound accuracy of the original model.}
\label{tab:hybrid_results_table}
\end{table*}

\Cref{tab:increasing_results_table} conducts the same experiments for the increasing schedule.
Interestingly, \textit{across all models and datasets}, token merging is outperforming other reduction strategies and random merging remains the worst strategy.
The action and noun accuracies on ViViT are actually \textit{improved} a small amount by token merging, which suggests that merging tokens well in the later layers might refine video features slightly.
Comparing the upper bound model accuracy with the merged counterparts, we see smaller drops in accuracy than in~\cref{tab:decreasing_results_table}.
For the divided space-time models, the accuracy on the temporally sensitive datasets (SSv$2$ and EK-$100$ verb accuracy) still demonstrate significant drops, showing that even when biasing merging towards the later layers, the models' ability to fuse temporal information is being hindered.
ViViT and VideoMAE demonstrate drops in accuracy of \textit{no more than $2\%$ across all datasets}, while gaining a speedup of roughly $1.6$X, indicating that an increasing schedule is an extremely ``safe'' option for these models, minimising accuracy trade-off.
We have demonstrated that with an increasing schedule and a reasonable $r$ value, token merging is the reduction strategy that preserves accuracy the best.

\subsection{Hybrid Merging}
\label{sub:hybrid_merging}

The final merged tokens are typically visually distinct clusters of image patches, a characteristic that can be observed in many qualitative examples in~\cref{sec:qualitative_examples}.
One obvious limitation of the token merging~\cite{bolya2022token} scheme is that it isn't adaptive in the sense that tokens are \textit{forced} to merge if their pairwise similarity is one of the $r$ largest.
Towards the tail end of the merging process, different tokens become more visually dissimilar, as the clusters become saturated.
We assume that merging these dissimilar tokens is destructive for performance, which we derive from the fact that random dropout is much preferable to random merging in~\cref{sub:quantitative_results}.

To determine whether this phenomenon presents itself in vanilla token merging and attempt to alleviate it, we experiment with a hybrid scheme of dropout and merging.
We a define a threshold $t$, where a token in the top $r$ pairs is \textit{dropped} instead of merged if the similarity is lower than $t$.
Using this strategy, the model can adaptively merge/drop tokens, ensuring that merging only happens when token pairs are similar past a set threshold.
In~\cref{tab:hybrid_results_table}, we have the results of an experiment applying hybrid merging to ViViT and VideoMAE, after ablating the threshold against EK-$100$ to find an optimal value.
Generally, these results highlight very similar performance to vanilla token merging, with the models showing consistent but small improvements on K$400$ and EK-$100$.
Though marginal, these results indicate that it may be possible to develop stronger representations of token sequences with a combination of both dropout and merging.

\subsection{ViViT Confusion Matrices}
\label{sub:vivit_confusion}

\begin{figure*}[htp]
  \centering
  \includegraphics[width=0.97\linewidth]{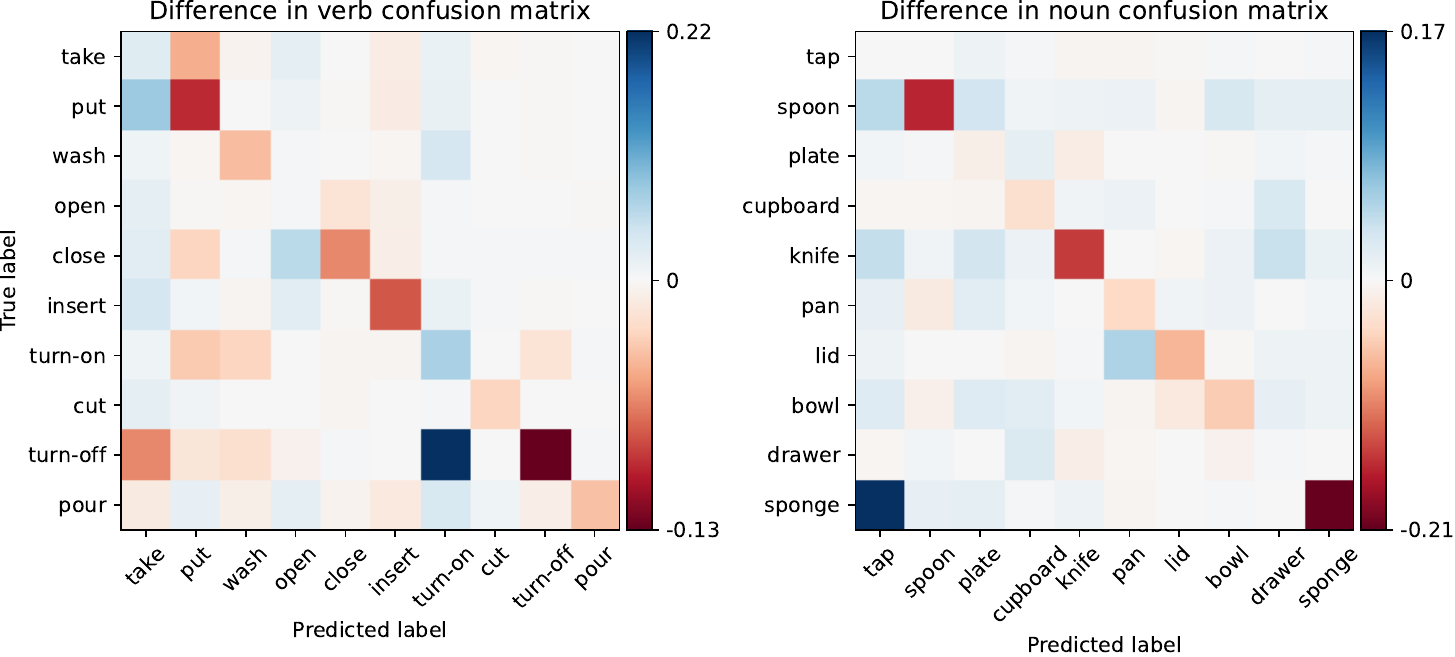}
  \caption{Impact on confusion matrices from Token Merging a ViViT model. The first $10$ verb and noun classes are displayed left and right respectively from ViViT on EK-$100$. Red indicates less predictions and blue indicates more predictions.}
  \label{fig:vivit_confusion_difference}
\end{figure*}

To investigate the errors introduced by merging with ViViT, we plot confusion matrices for the most frequent $10$ verb and noun classes in EK-$100$. 
In~\cref{fig:vivit_confusion_difference} we have the difference in performance between a model \textit{with} and a model \textit{without} token merging. 
We can see a somewhat similar trend to the same figure produced for VideoMAE, with less predictions being introduced in the diagonal and more predictions elsewhere, though the model is clearly more resilient to confusion. 
For the verbs in~\cref{fig:vivit_confusion_difference} (left), we see that the model does not collapse towards the ``take'' and ``put'' classes.
In fact, the largest change in performance sees ``turn-off'' being misclassified as ``turn-on''.
These verbs are essentially the same action with slightly different context, suggesting that merging causes confusion among visually similar actions. 
Most of the nouns in~\cref{fig:vivit_confusion_difference} (right) do not shift significantly, however small objects like ``spoon'', ``knife'' and ``sponge'' are increasingly misclassified. As well as this, ``sponge'' is confused with ``tap'' particularly often. This again suggests that the features of small objects are the first to be lost by merging tokens.

\section{Training Hyperparameters}
\label{sec:training_hyperparameters}

\begin{table*}[htp]
\centering
\begin{tabular}{cccc}
\hline
 & TimeSformer & VideoMAE & ViViT \\ \hline
Batch size & 128 & 128 & 64 \\
Gradient accumulation steps & 1 & 1 & 1 \\
Base learning rate & 5e-3 & 1e-3 & 1e-2 \\
Learning rate policy & Step with relative LRs & Cosine with cosine warmup & Cosine with cosine warmup \\
Warmup learning rate & - & 0 & 0 \\
Warmup epochs & - & 5.0 & 2.5 \\
Epochs & 15 & 50 & 50 \\
Optimiser & SGD & AdamW & SGD \\
Momentum & 0.9 & 0.9 & 0.9 \\ \hline
\end{tabular}
\caption{Hyperparameters used to train TimeSformer, VideoMAE and ViViT checkpoints with four GH$200$ GPUs~\cite{mcintoshsmith2024}, used to evaluate merging on EK-$100$. Where possible we tried to reproduce the setup used in the original works.}
\label{tab:training_hyperparameters}
\end{table*}

As we've mentioned in the main paper, due to the fact that TimeSformer, VideoMAE and ViViT did not have freely available checkpoints for EK-$100$ online, we were required to finetune our own checkpoints for evaluation.
In~\cref{tab:training_hyperparameters}, we have an overview of the hyperparameters we used.
These have been adapted (with as few changes as possible) from \cite{bertasius2021space,tong2022videomae,arnab2021vivit}.
For TimeSformer, the learning rate is multiplied by $1$, $0.1$ and $0.01$, at the first, twelfth and last epochs respectively.
We do not use Mixup~\cite{zhang2018mixup} when finetuning.

\section{Qualitative Examples}
\label{sec:qualitative_examples}

Here we collect a range of qualitative examples that further our claims made in the main paper.
To gather these, we generated visualisations at random and then kept a mixture of simple cases where (video subjects can be easily tracked across all frames) and more complex cases (where motion blur, occlusions or totally new subjects are introduced mid video).

\begin{figure*}[htp]
  \centering
  \begin{subfigure}{1.0\linewidth}
    \centering
    \includegraphics[width=1.0\linewidth]{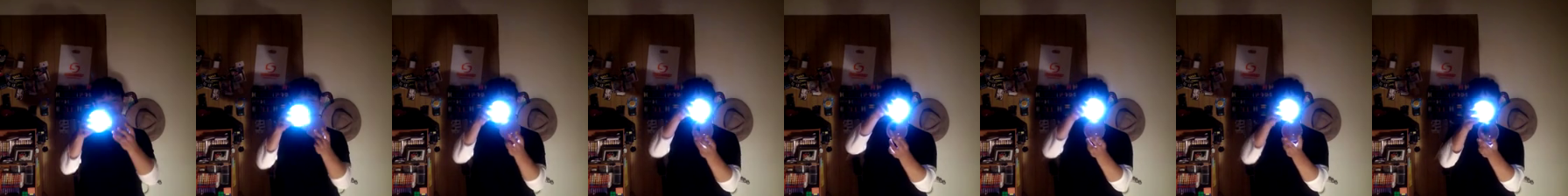}
    \caption{Original clip.}
    \label{fig:kinetics_example_a}
  \end{subfigure}
  \hfill
  \begin{subfigure}{1.0\linewidth}
    \centering
    \includegraphics[width=1.0\linewidth]{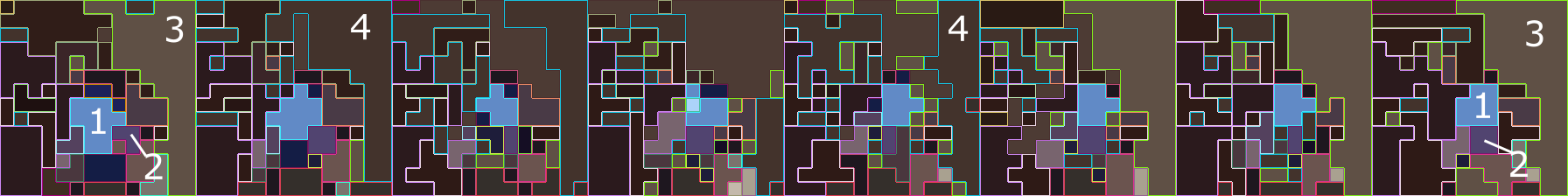}
    \caption{Merged clip.}
    \label{fig:kinetics_example_b}
  \end{subfigure}
  \caption{Visualisation of the final merged tokens for a K$400$ clip of ``contact juggling'', produced with VideoMAE. Token $1$ tracks the bright blue ball, while token $2$ tracks the darker ball rotating it.}
  \label{fig:kinetics_example}
\end{figure*}

First, we include a visualisation of tokens merging through a K$400$ clip in~\cref{fig:kinetics_example}, where tokens of interest have been numbered like in the main paper. 
In~\cref{fig:kinetics_example_b}, we have the final merged tokens for an example of ``contact juggling''.
The bright blue ball is well captured across the entire clip by token $1$, with the darker blue ball being captured by token $2$.
Notably, token $2$ is tracked rotating clockwise around token $1$.
Tokens $3$ and $4$ represent the wall behind the right of the person, though the left side of the wall is not well merged, likely due to there being no common texture.
In this case, video token merging is capable of tracking visually similar objects through all frames of the clip.

\begin{figure*}[htp]
  \centering
  \begin{subfigure}{0.96\linewidth}
    \centering
    \includegraphics[width=1.0\linewidth]{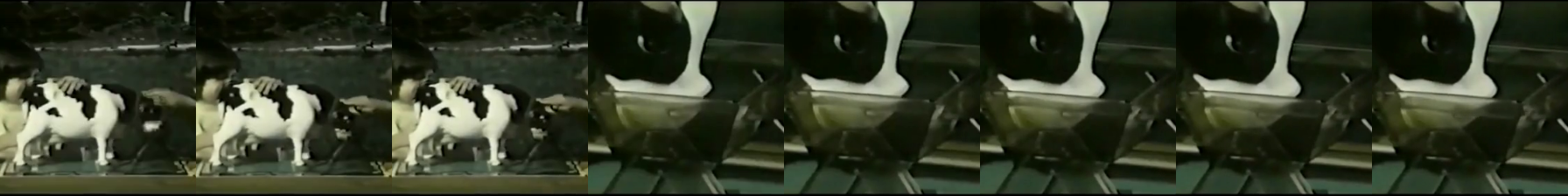}
  \end{subfigure}
  \begin{subfigure}{0.96\linewidth}
    \centering
    \includegraphics[width=1.0\linewidth]{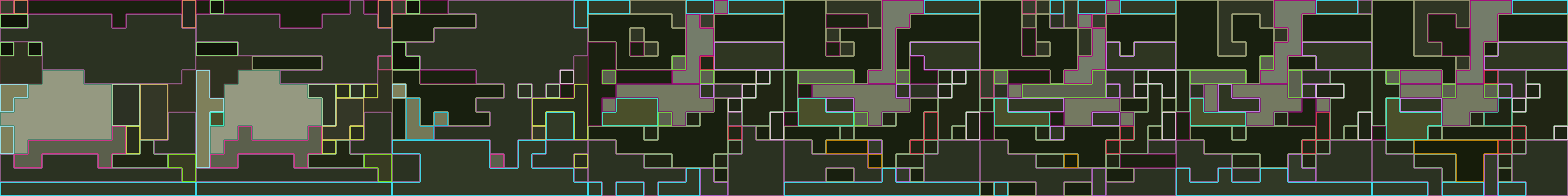}
  \end{subfigure}
  \hfill
  \begin{subfigure}{0.96\linewidth}
    \centering
    \includegraphics[width=1.0\linewidth]{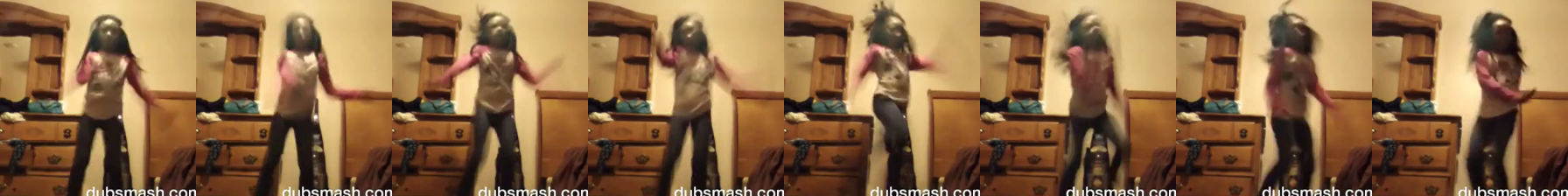}
  \end{subfigure}
  \begin{subfigure}{0.96\linewidth}
    \centering
    \includegraphics[width=1.0\linewidth]{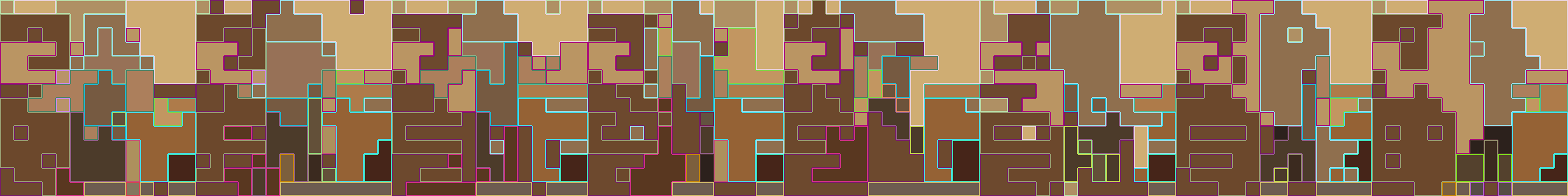}
  \end{subfigure}
  \hfill
  \begin{subfigure}{0.96\linewidth}
    \centering
    \includegraphics[width=1.0\linewidth]{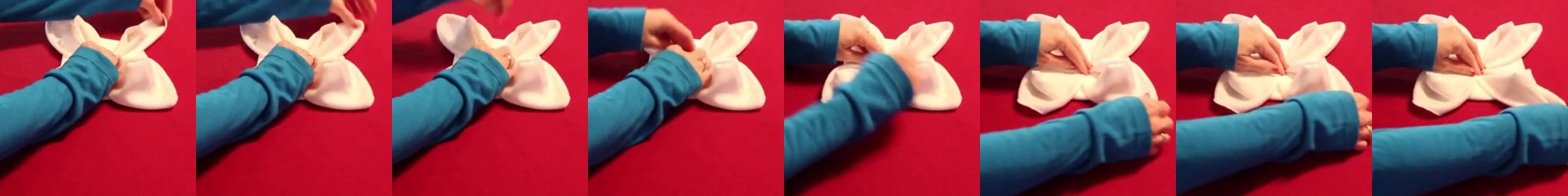}
  \end{subfigure}
  \begin{subfigure}{0.96\linewidth}
    \centering
    \includegraphics[width=1.0\linewidth]{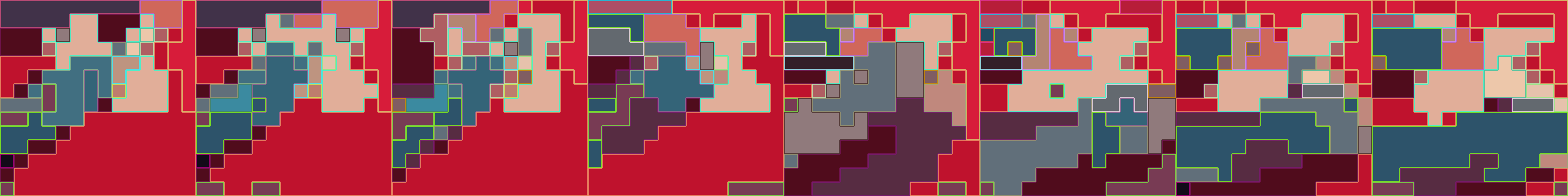}
  \end{subfigure}
  \hfill
  \begin{subfigure}{0.96\linewidth}
    \centering
    \includegraphics[width=1.0\linewidth]{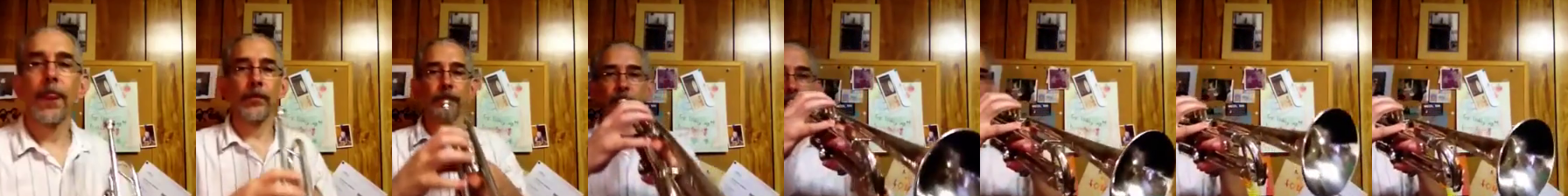}
  \end{subfigure}
  \begin{subfigure}{0.96\linewidth}
    \centering
    \includegraphics[width=1.0\linewidth]{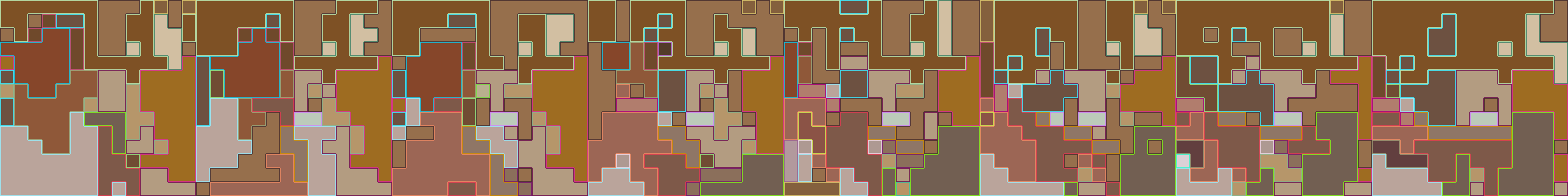}
  \end{subfigure}
  \hfill
  \begin{subfigure}{0.96\linewidth}
    \centering
    \includegraphics[width=1.0\linewidth]{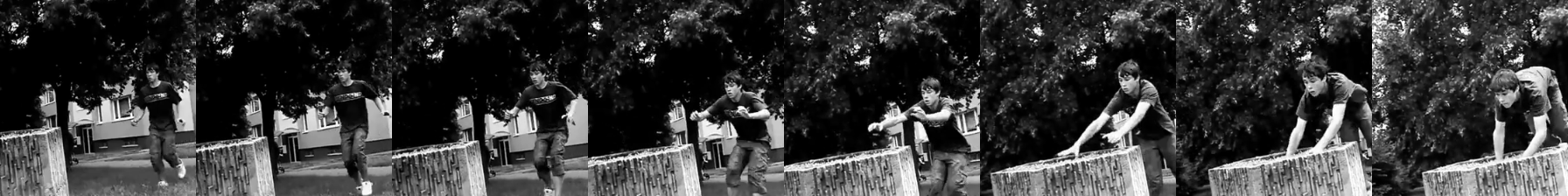}
  \end{subfigure}
  \begin{subfigure}{0.96\linewidth}
    \centering
    \includegraphics[width=1.0\linewidth]{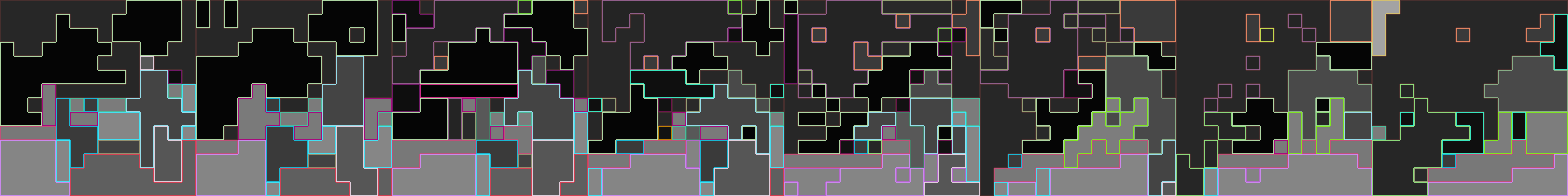}
  \end{subfigure}
  \caption{Visualisations of the final merged tokens for K400 clips, produced with VideoMAE.}
  \label{fig:videomae_kinetics_examples}
\end{figure*}

\begin{figure*}[htp]
  \centering
  \begin{subfigure}{0.96\linewidth}
    \centering
    \includegraphics[width=1.0\linewidth]{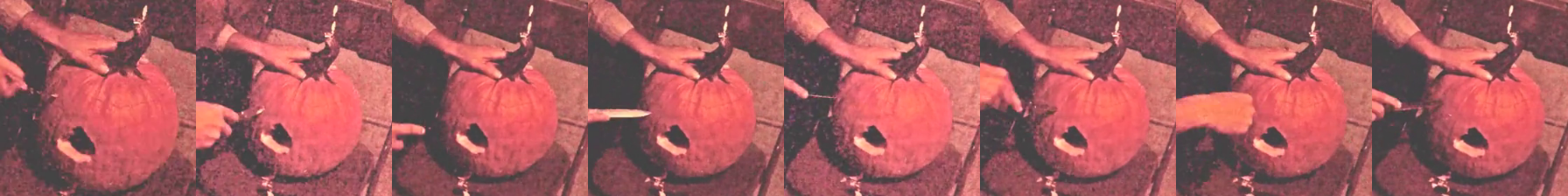}
  \end{subfigure}
  \begin{subfigure}{0.96\linewidth}
    \centering
    \includegraphics[width=1.0\linewidth]{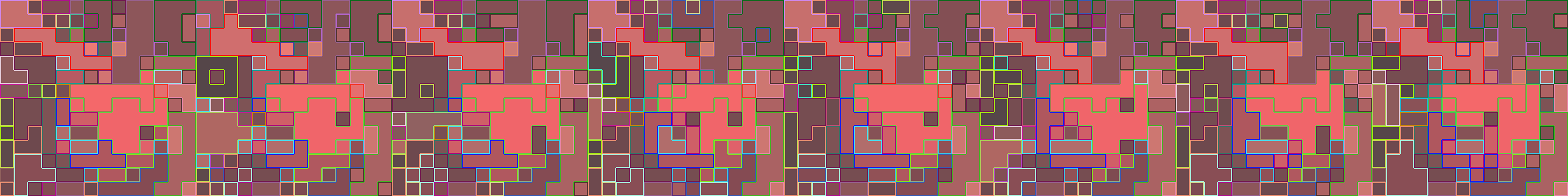}
  \end{subfigure}
  \hfill
  \begin{subfigure}{0.96\linewidth}
    \centering
    \includegraphics[width=1.0\linewidth]{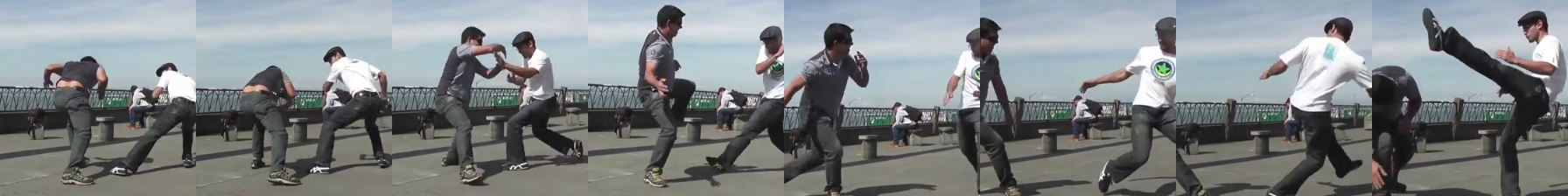}
  \end{subfigure}
  \begin{subfigure}{0.96\linewidth}
    \centering
    \includegraphics[width=1.0\linewidth]{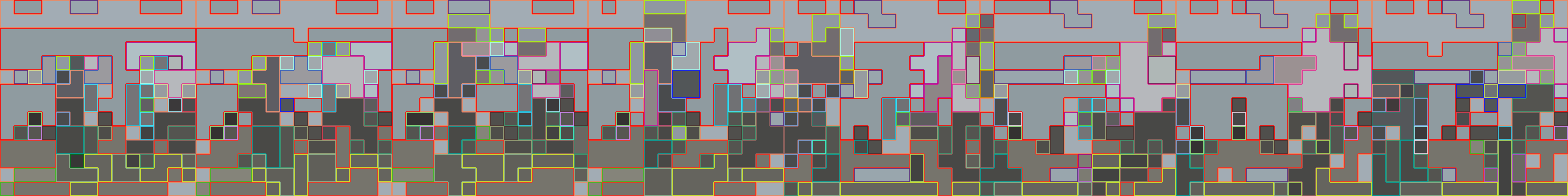}
  \end{subfigure}
  \hfill
  \begin{subfigure}{0.96\linewidth}
    \centering
    \includegraphics[width=1.0\linewidth]{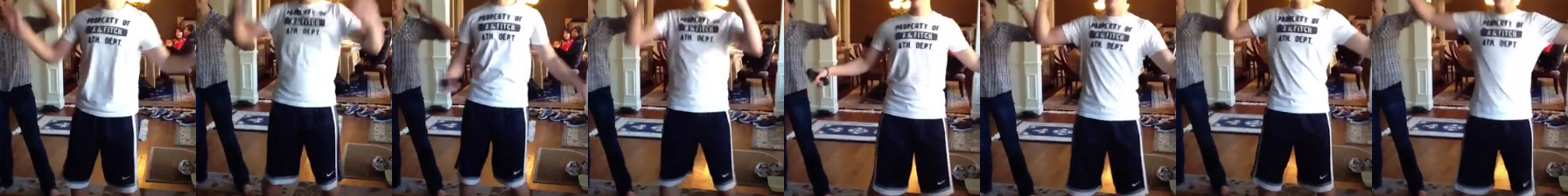}
  \end{subfigure}
  \begin{subfigure}{0.96\linewidth}
    \centering
    \includegraphics[width=1.0\linewidth]{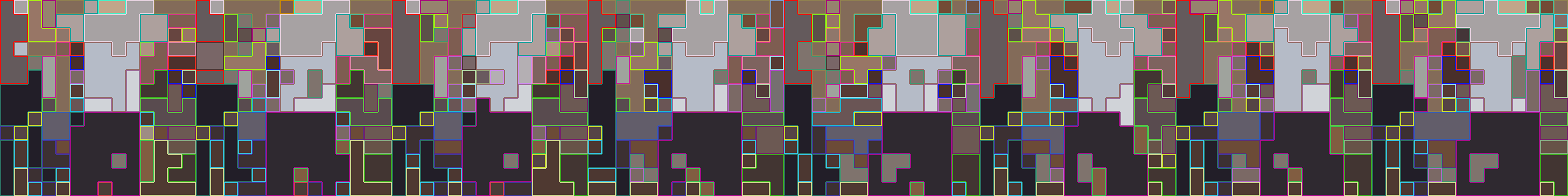}
  \end{subfigure}
  \hfill
  \begin{subfigure}{0.96\linewidth}
    \centering
    \includegraphics[width=1.0\linewidth]{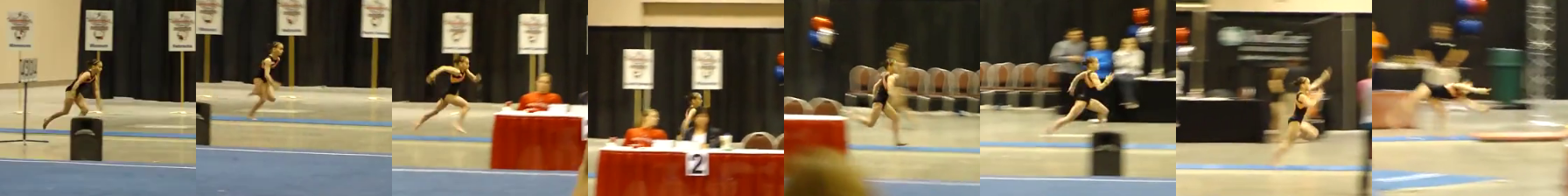}
  \end{subfigure}
  \begin{subfigure}{0.96\linewidth}
    \centering
    \includegraphics[width=1.0\linewidth]{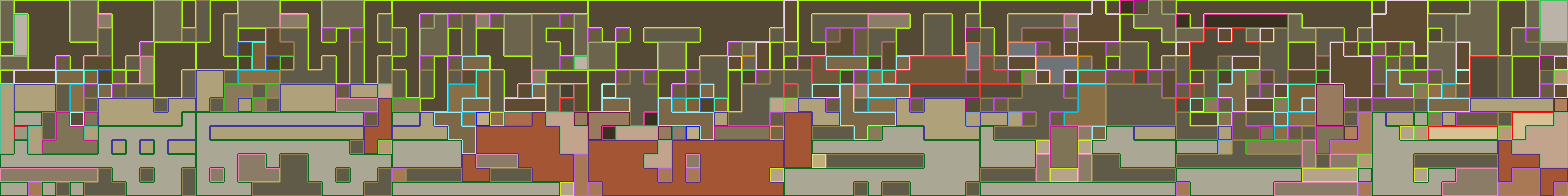}
  \end{subfigure}
  \hfill
  \begin{subfigure}{0.96\linewidth}
    \centering
    \includegraphics[width=1.0\linewidth]{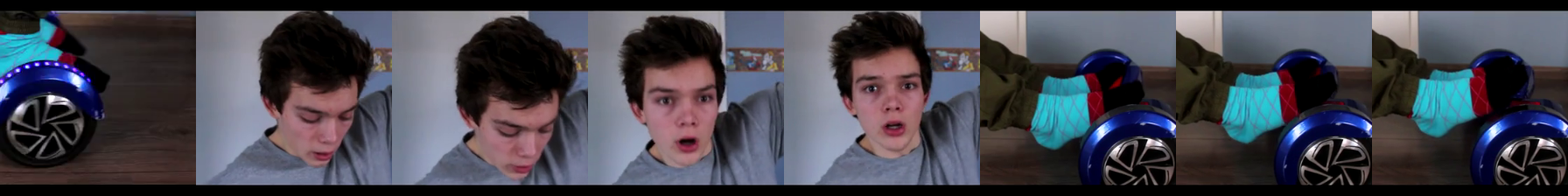}
  \end{subfigure}
  \begin{subfigure}{0.96\linewidth}
    \centering
    \includegraphics[width=1.0\linewidth]{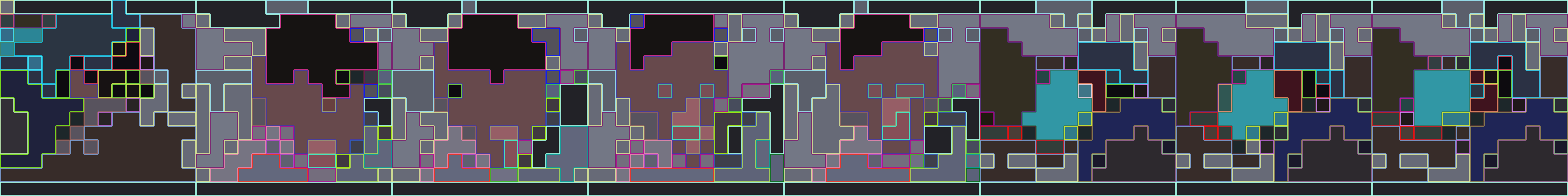}
  \end{subfigure}
  \caption{Visualisations of the final merged tokens for K400 clips, produced with ViViT.}
  \label{fig:vivit_kinetics_examples}
\end{figure*}

\begin{figure*}[htp]
  \centering
  \begin{subfigure}{0.96\linewidth}
    \centering
    \includegraphics[width=1.0\linewidth]{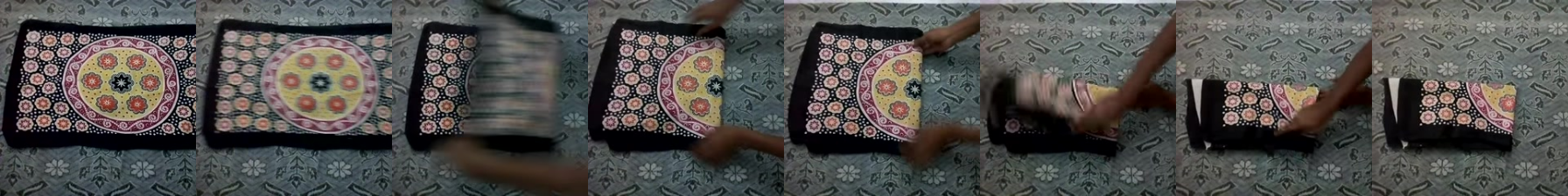}
  \end{subfigure}
  \begin{subfigure}{0.96\linewidth}
    \centering
    \includegraphics[width=1.0\linewidth]{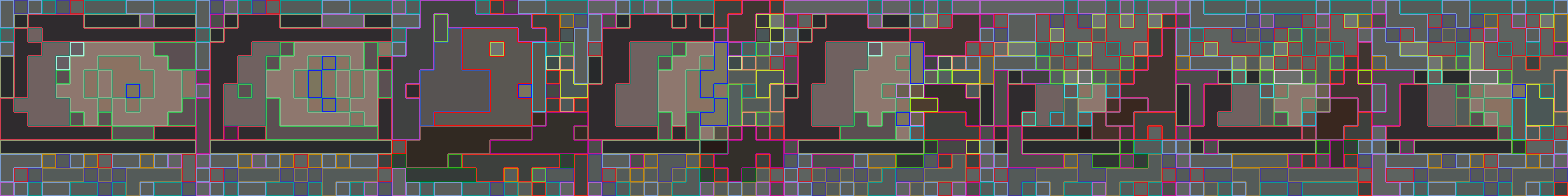}
  \end{subfigure}
  \hfill
  \begin{subfigure}{0.96\linewidth}
    \centering
    \includegraphics[width=1.0\linewidth]{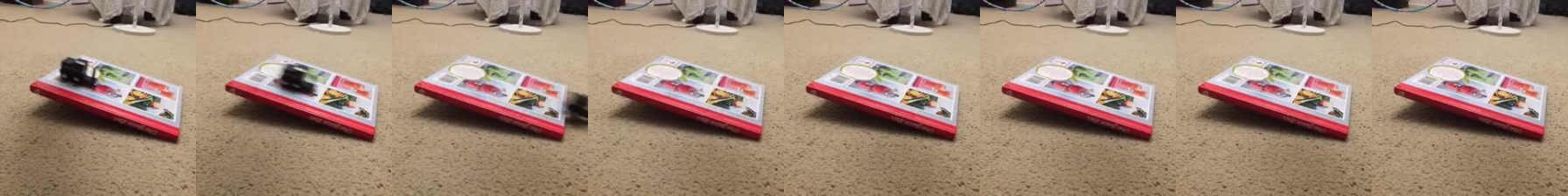}
  \end{subfigure}
  \begin{subfigure}{0.96\linewidth}
    \centering
    \includegraphics[width=1.0\linewidth]{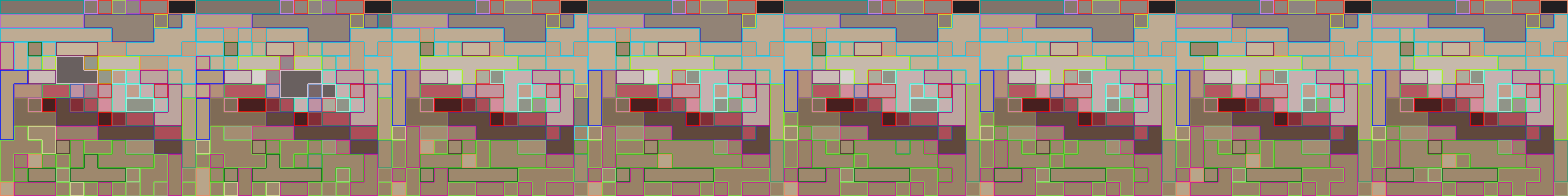}
  \end{subfigure}
  \hfill
  \begin{subfigure}{0.96\linewidth}
    \centering
    \includegraphics[width=1.0\linewidth]{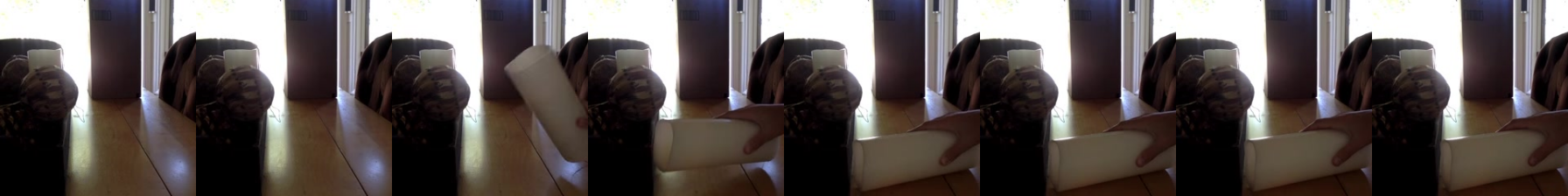}
  \end{subfigure}
  \begin{subfigure}{0.96\linewidth}
    \centering
    \includegraphics[width=1.0\linewidth]{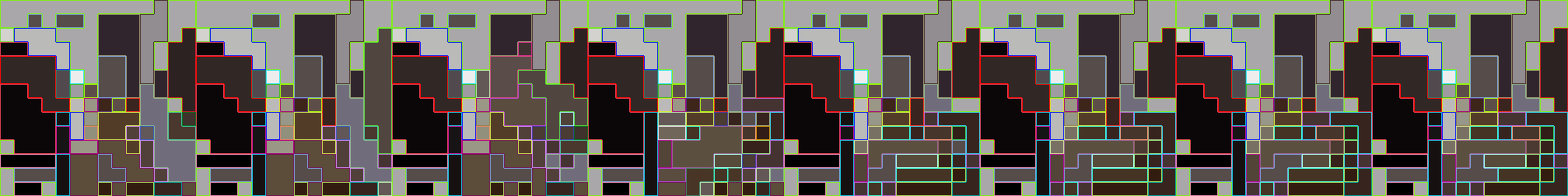}
  \end{subfigure}
  \hfill
  \begin{subfigure}{0.96\linewidth}
    \centering
    \includegraphics[width=1.0\linewidth]{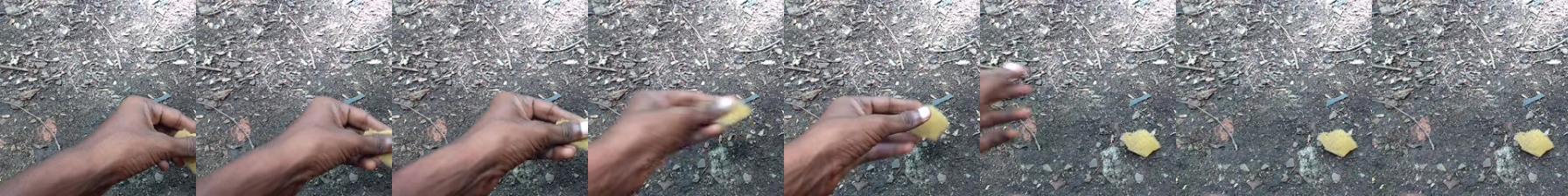}
  \end{subfigure}
  \begin{subfigure}{0.96\linewidth}
    \centering
    \includegraphics[width=1.0\linewidth]{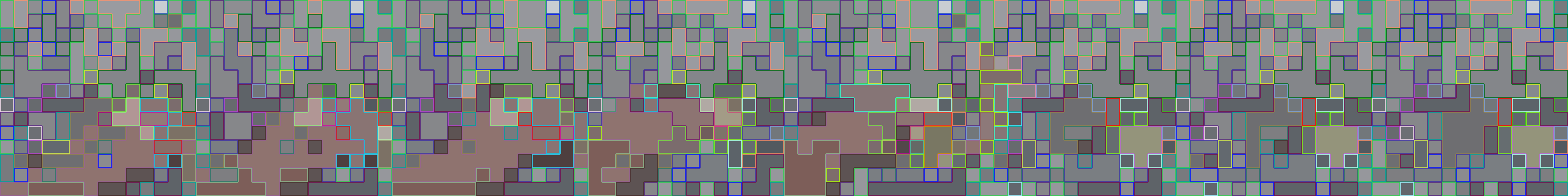}
  \end{subfigure}
  \hfill
  \begin{subfigure}{0.96\linewidth}
    \centering
    \includegraphics[width=1.0\linewidth]{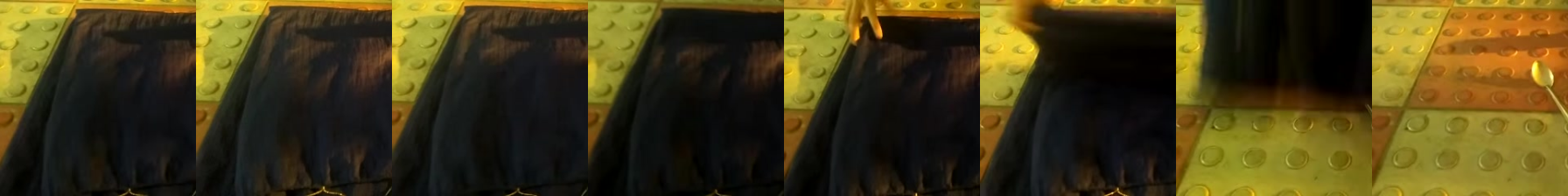}
  \end{subfigure}
  \begin{subfigure}{0.96\linewidth}
    \centering
    \includegraphics[width=1.0\linewidth]{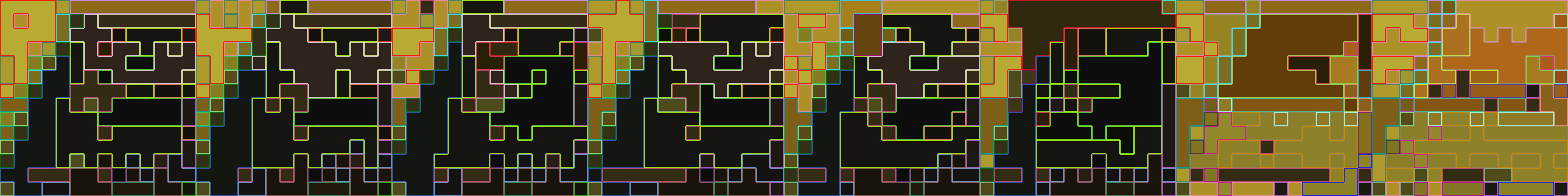}
  \end{subfigure}
  \caption{Visualisations of the final merged tokens for SSv2 clips, produced with ViViT.}
  \label{fig:vivit_ssv2_examples}
\end{figure*}

\begin{figure*}[htp]
  \centering
  \begin{subfigure}{0.96\linewidth}
    \centering
    \includegraphics[width=1.0\linewidth]{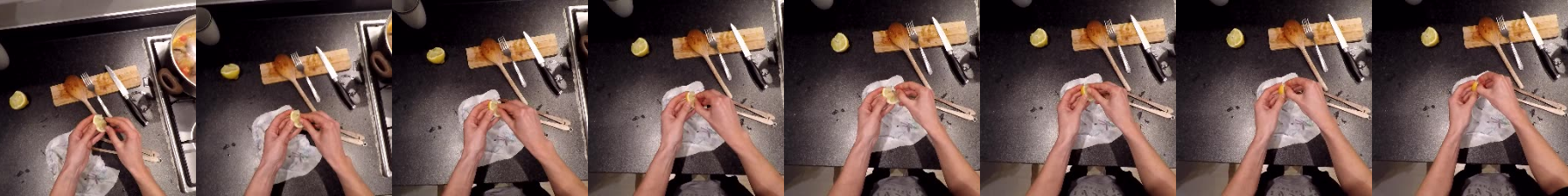}
  \end{subfigure}
  \begin{subfigure}{0.96\linewidth}
    \centering
    \includegraphics[width=1.0\linewidth]{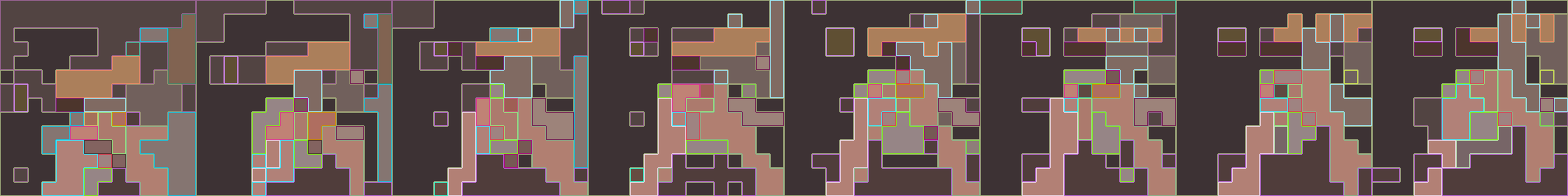}
  \end{subfigure}
  \hfill
  \begin{subfigure}{0.96\linewidth}
    \centering
    \includegraphics[width=1.0\linewidth]{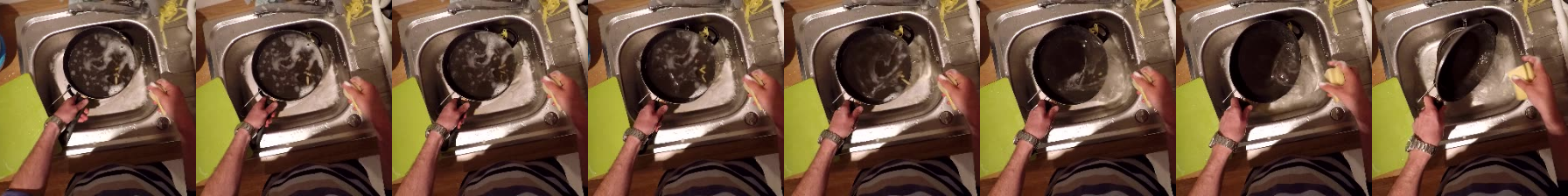}
  \end{subfigure}
  \begin{subfigure}{0.96\linewidth}
    \centering
    \includegraphics[width=1.0\linewidth]{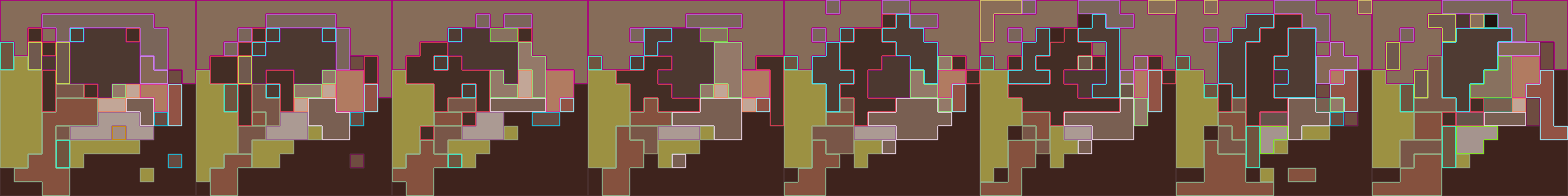}
  \end{subfigure}
  \hfill
  \begin{subfigure}{0.96\linewidth}
    \centering
    \includegraphics[width=1.0\linewidth]{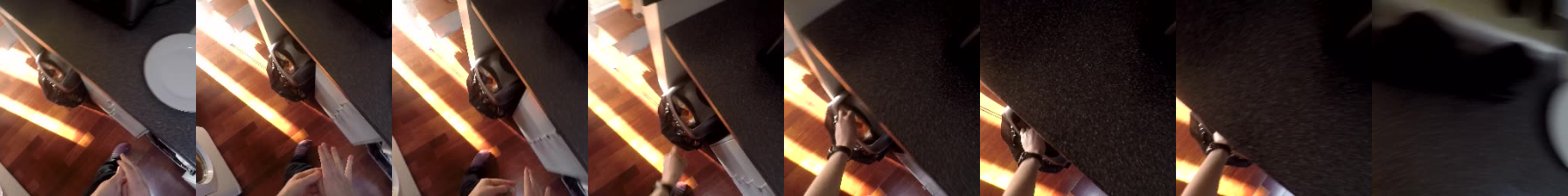}
  \end{subfigure}
  \begin{subfigure}{0.96\linewidth}
    \centering
    \includegraphics[width=1.0\linewidth]{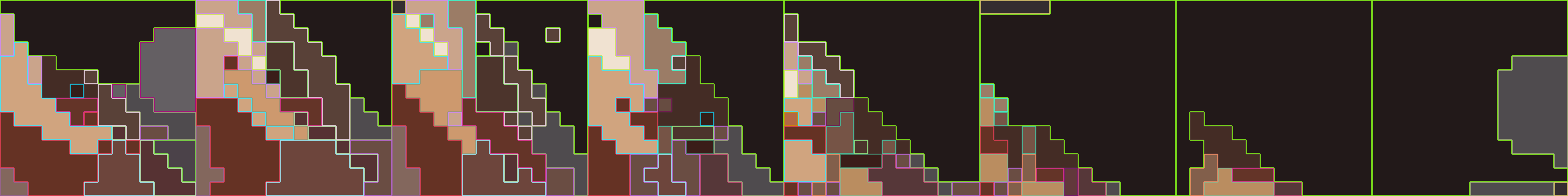}
  \end{subfigure}
  \hfill
  \begin{subfigure}{0.96\linewidth}
    \centering
    \includegraphics[width=1.0\linewidth]{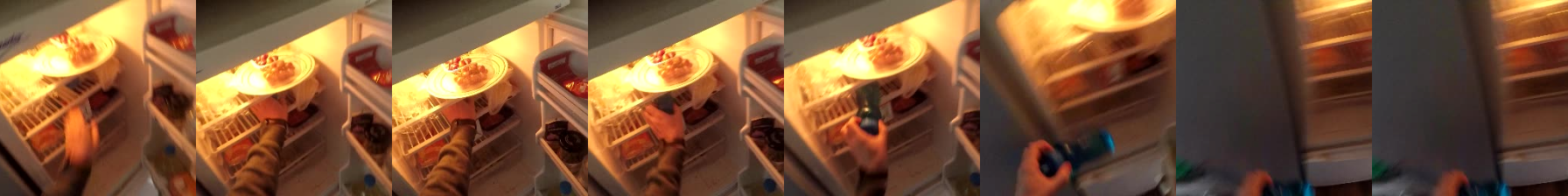}
  \end{subfigure}
  \begin{subfigure}{0.96\linewidth}
    \centering
    \includegraphics[width=1.0\linewidth]{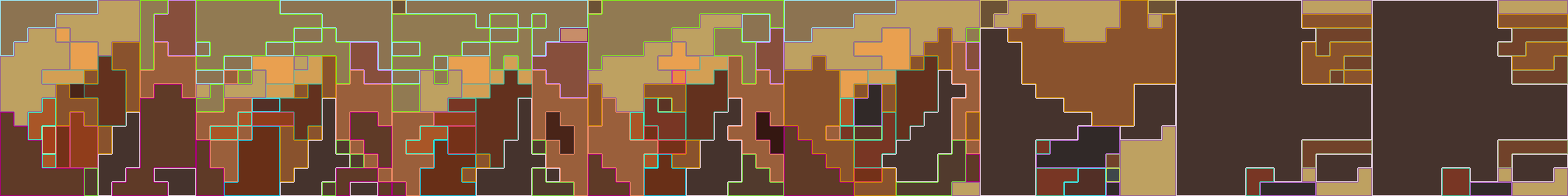}
  \end{subfigure}
  \hfill
  \begin{subfigure}{0.96\linewidth}
    \centering
    \includegraphics[width=1.0\linewidth]{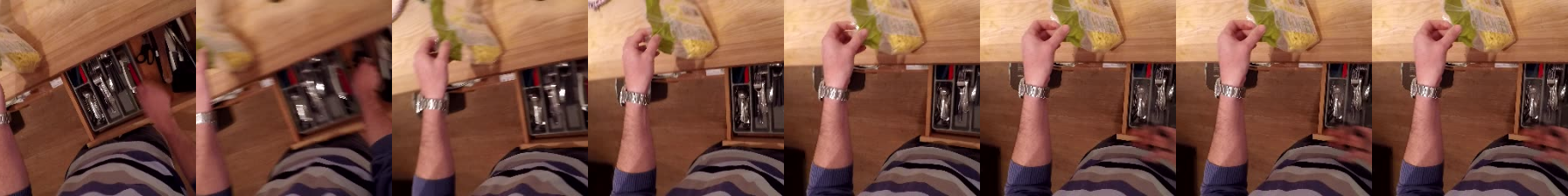}
  \end{subfigure}
  \begin{subfigure}{0.96\linewidth}
    \centering
    \includegraphics[width=1.0\linewidth]{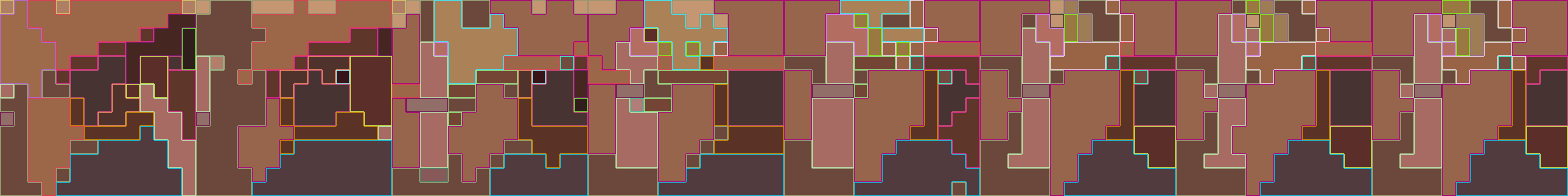}
  \end{subfigure}
  \caption{Visualisations of the final merged tokens for EK-100 clips, produced with VideoMAE.}
  \label{fig:videomae_epickitchens_examples}
\end{figure*}

\begin{figure*}[htp]
  \centering
  \begin{subfigure}{0.96\linewidth}
    \centering
    \includegraphics[width=1.0\linewidth]{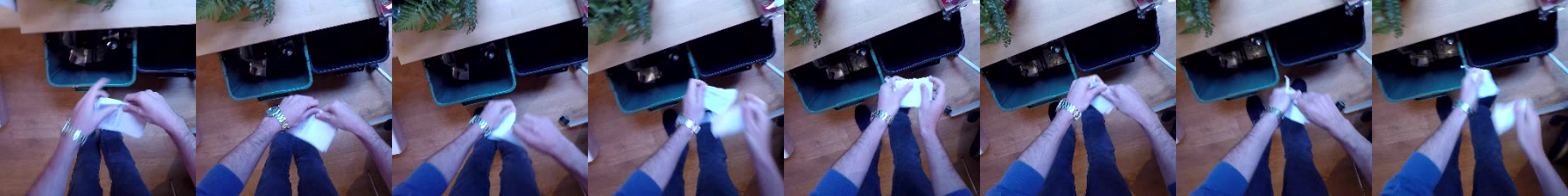}
  \end{subfigure}
  \begin{subfigure}{0.96\linewidth}
    \centering
    \includegraphics[width=1.0\linewidth]{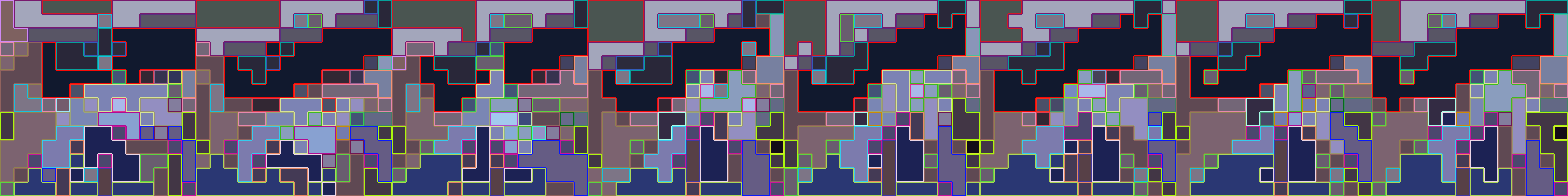}
  \end{subfigure}
  \hfill
  \begin{subfigure}{0.96\linewidth}
    \centering
    \includegraphics[width=1.0\linewidth]{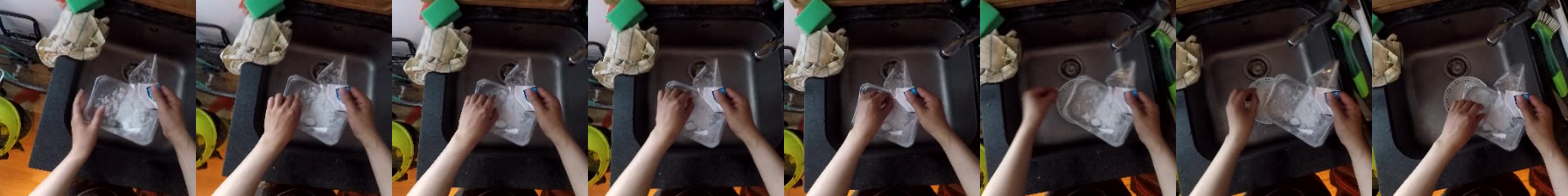}
  \end{subfigure}
  \begin{subfigure}{0.96\linewidth}
    \centering
    \includegraphics[width=1.0\linewidth]{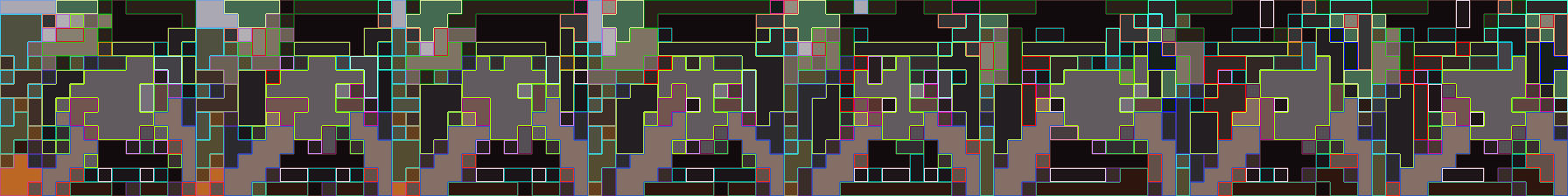}
  \end{subfigure}
  \hfill
  \begin{subfigure}{0.96\linewidth}
    \centering
    \includegraphics[width=1.0\linewidth]{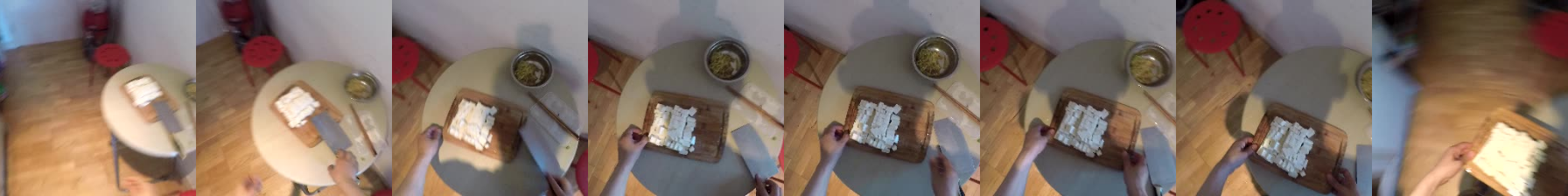}
  \end{subfigure}
  \begin{subfigure}{0.96\linewidth}
    \centering
    \includegraphics[width=1.0\linewidth]{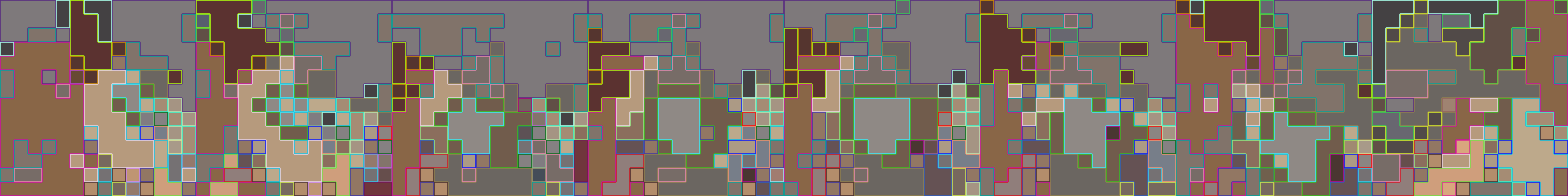}
  \end{subfigure}
  \hfill
  \begin{subfigure}{0.96\linewidth}
    \centering
    \includegraphics[width=1.0\linewidth]{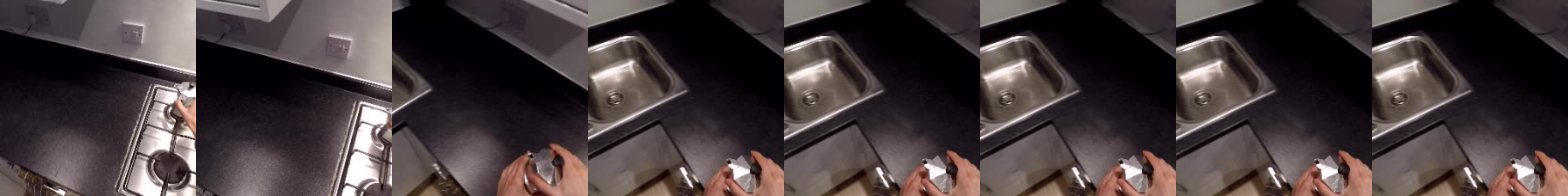}
  \end{subfigure}
  \begin{subfigure}{0.96\linewidth}
    \centering
    \includegraphics[width=1.0\linewidth]{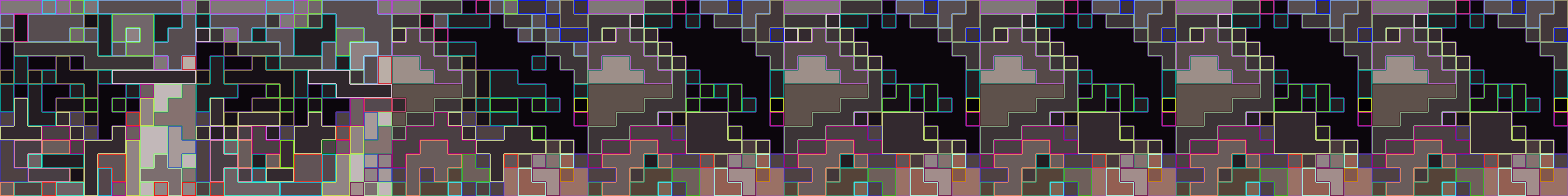}
  \end{subfigure}
  \hfill
  \begin{subfigure}{0.96\linewidth}
    \centering
    \includegraphics[width=1.0\linewidth]{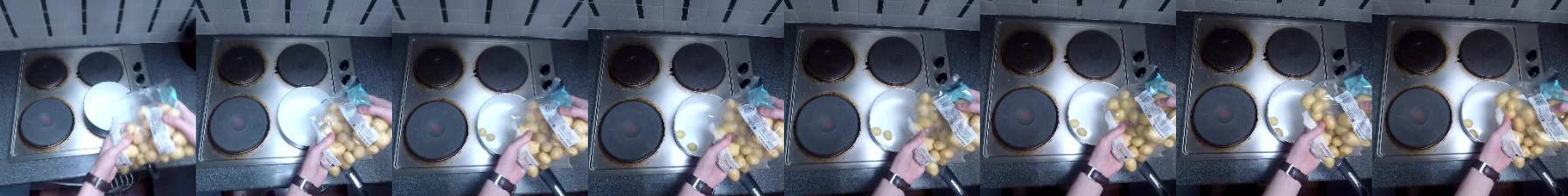}
  \end{subfigure}
  \begin{subfigure}{0.96\linewidth}
    \centering
    \includegraphics[width=1.0\linewidth]{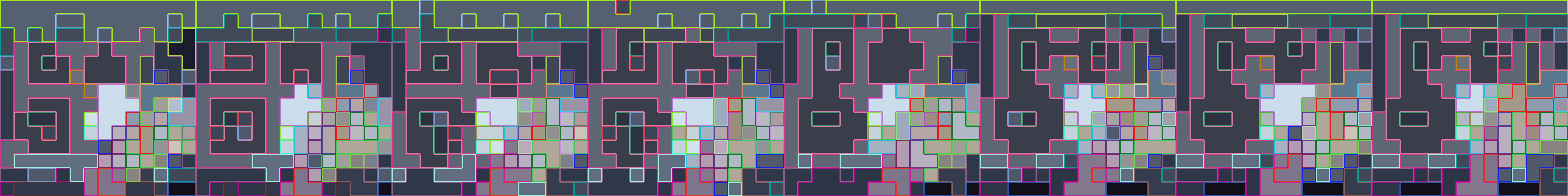}
  \end{subfigure}
  \caption{Visualisations of the final merged tokens for EK-100 clips, produced with ViViT.}
  \label{fig:vivit_epickitchens_examples}
\end{figure*}

Next, we explore merging visualisations generated by both VideoMAE and ViViT. 
Firstly, from~\cref{fig:videomae_kinetics_examples} to~\cref{fig:vivit_epickitchens_examples} we present visualisations of final merged tokens for K$400$, SSv$2$ and EK-$100$ respectively. 
Interestingly, in EK-$100$ examples exhibiting lots of head movement and motion blur, VideoMAE appears to handle these cases better by producing clearer segmentations of the clip.
When hands are present near the centre of the frame, both models are capable of differentiating this from the objects the participant is interacting with.

\begin{figure*}[htp]
  \centering
  \begin{subfigure}{0.96\linewidth}
    \centering
    \includegraphics[width=1.0\linewidth]{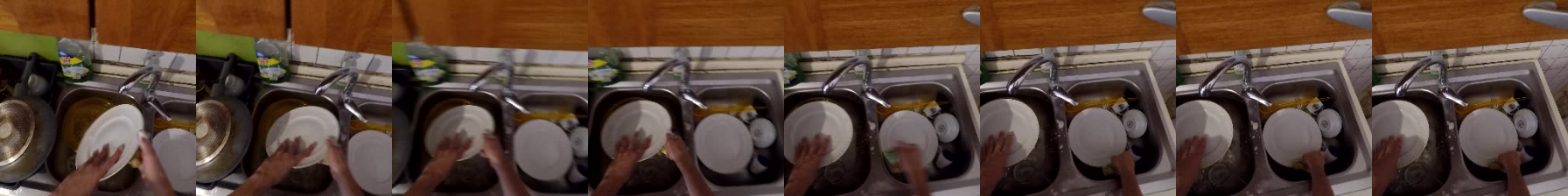}
  \end{subfigure}
  \begin{subfigure}{0.96\linewidth}
    \centering
    \includegraphics[width=1.0\linewidth]{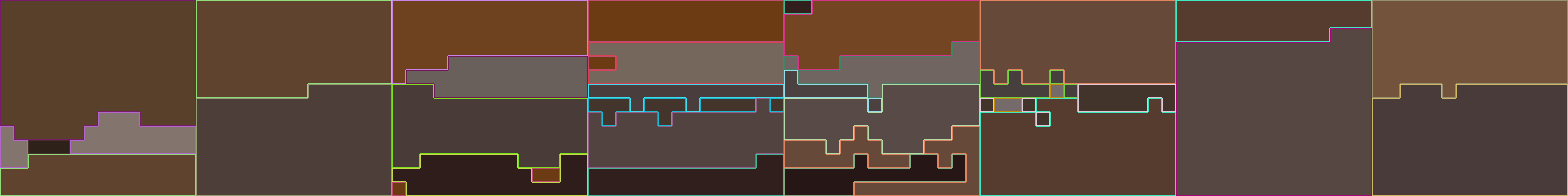}
  \end{subfigure}
  \begin{subfigure}{0.96\linewidth}
    \centering
    \includegraphics[width=1.0\linewidth]{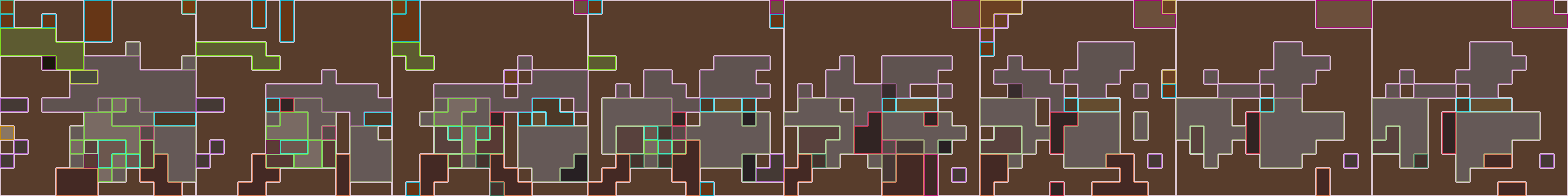}
  \end{subfigure}
  \hfill
  \begin{subfigure}{0.96\linewidth}
    \centering
    \includegraphics[width=1.0\linewidth]{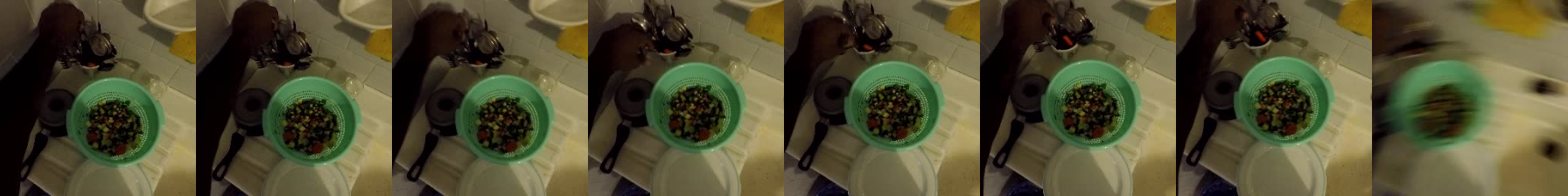}
  \end{subfigure}
  \begin{subfigure}{0.96\linewidth}
    \centering
    \includegraphics[width=1.0\linewidth]{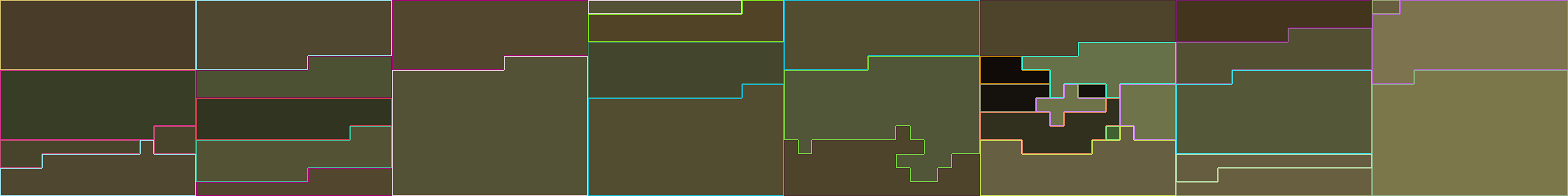}
  \end{subfigure}
  \begin{subfigure}{0.96\linewidth}
    \centering
    \includegraphics[width=1.0\linewidth]{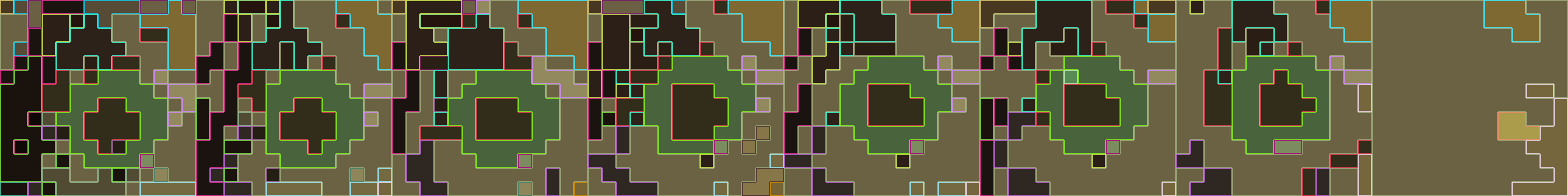}
  \end{subfigure}
  \hfill
  \begin{subfigure}{0.96\linewidth}
    \centering
    \includegraphics[width=1.0\linewidth]{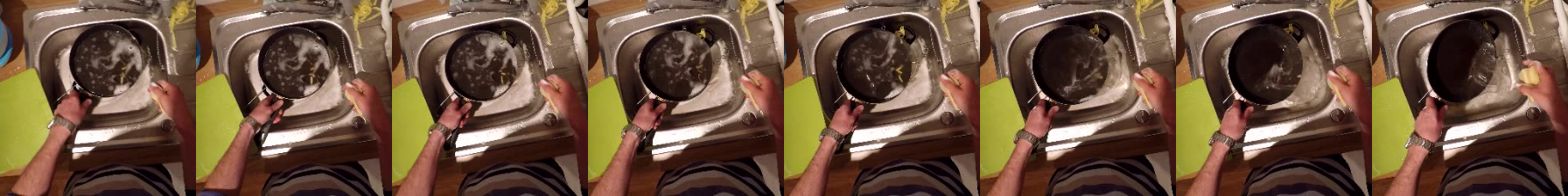}
  \end{subfigure}
  \begin{subfigure}{0.96\linewidth}
    \centering
    \includegraphics[width=1.0\linewidth]{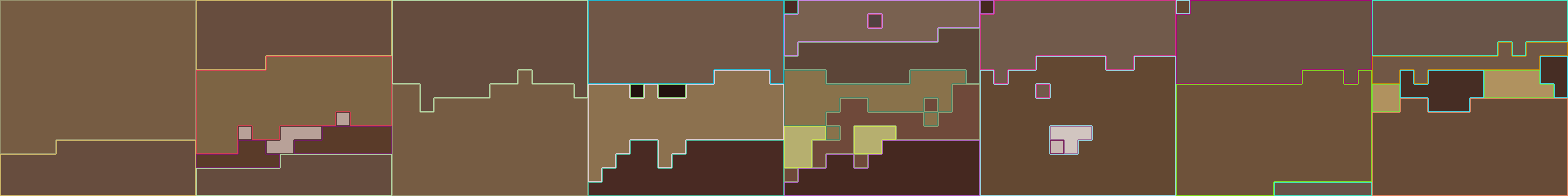}
  \end{subfigure}
  \begin{subfigure}{0.96\linewidth}
    \centering
    \includegraphics[width=1.0\linewidth]{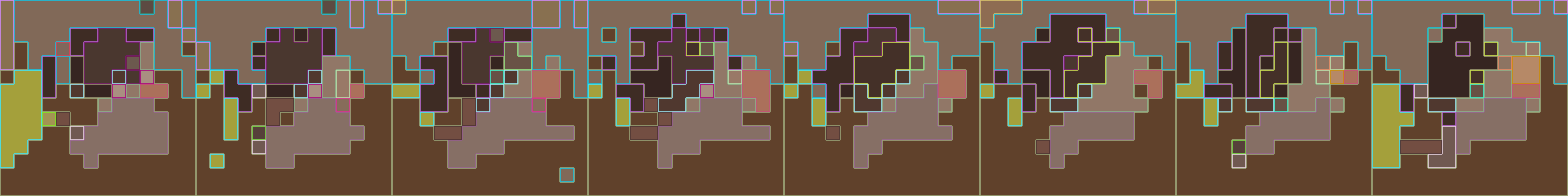}
  \end{subfigure}
  \caption{Visualisations of the difference in merging decisions made in layer $1$ versus layer $12$, produced with VideoMAE.}
  \label{fig:videomae_epickitchens_duplicate_layer_examples}
\end{figure*}

\begin{figure*}[htp]
  \centering
  \begin{subfigure}{0.96\linewidth}
    \centering
    \includegraphics[width=1.0\linewidth]{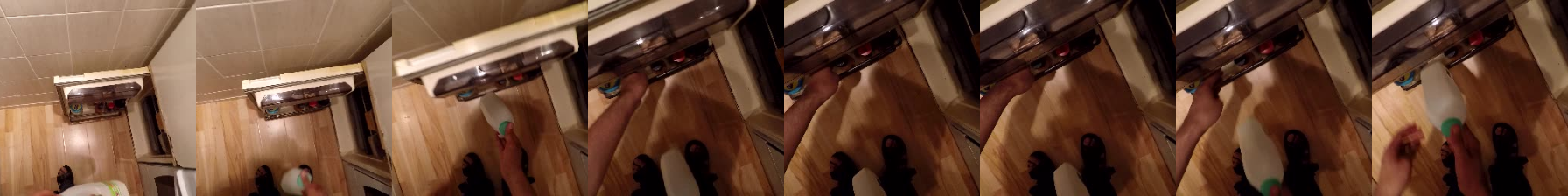}
  \end{subfigure}
  \begin{subfigure}{0.96\linewidth}
    \centering
    \includegraphics[width=1.0\linewidth]{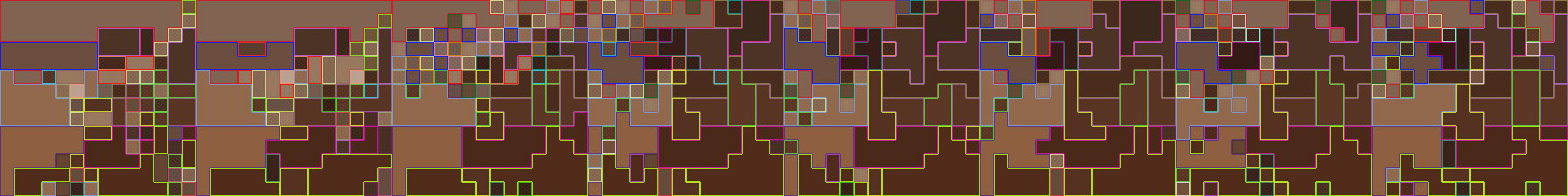}
  \end{subfigure}
  \begin{subfigure}{0.96\linewidth}
    \centering
    \includegraphics[width=1.0\linewidth]{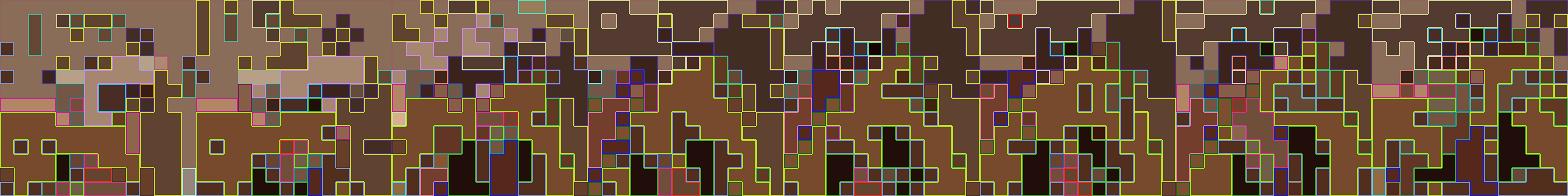}
  \end{subfigure}
  \hfill
  \begin{subfigure}{0.96\linewidth}
    \centering
    \includegraphics[width=1.0\linewidth]{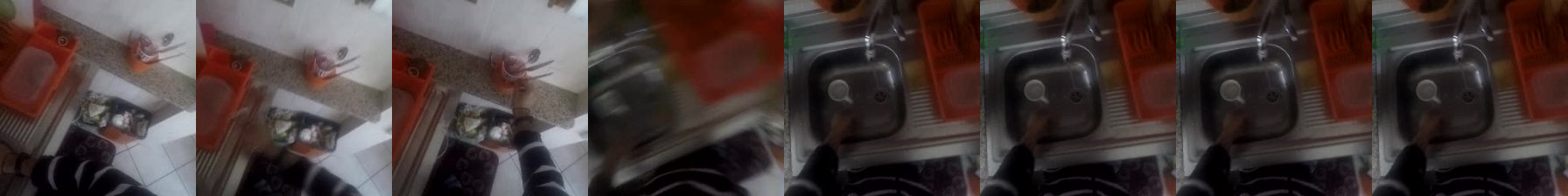}
  \end{subfigure}
  \begin{subfigure}{0.96\linewidth}
    \centering
    \includegraphics[width=1.0\linewidth]{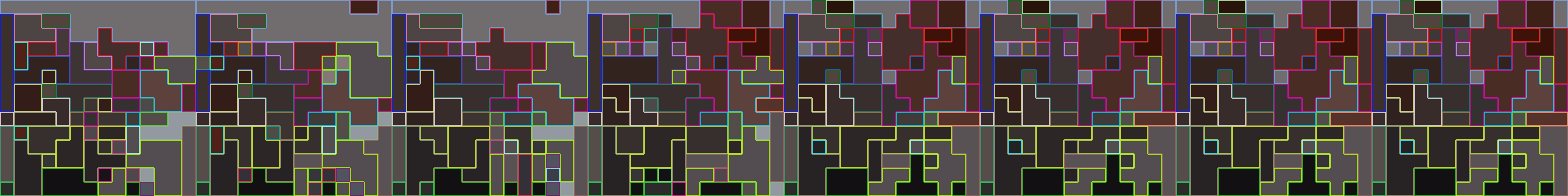}
  \end{subfigure}
  \begin{subfigure}{0.96\linewidth}
    \centering
    \includegraphics[width=1.0\linewidth]{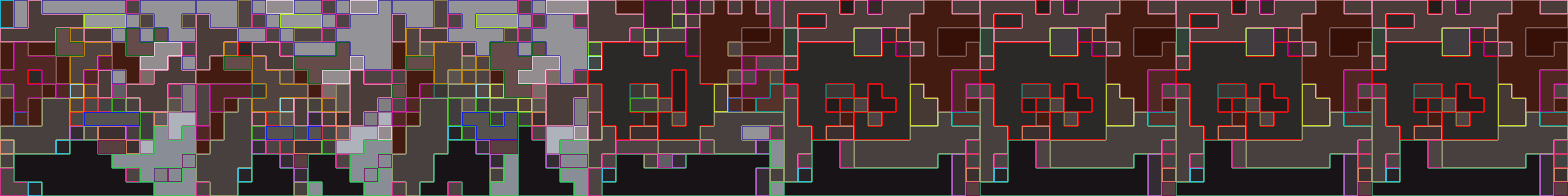}
  \end{subfigure}
  \hfill
  \begin{subfigure}{0.96\linewidth}
    \centering
    \includegraphics[width=1.0\linewidth]{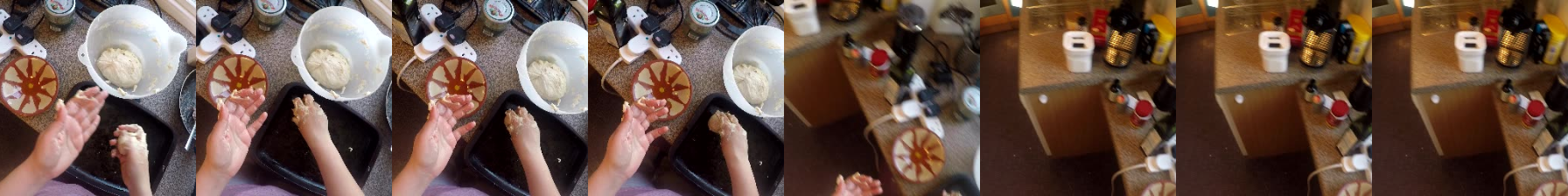}
  \end{subfigure}
  \begin{subfigure}{0.96\linewidth}
    \centering
    \includegraphics[width=1.0\linewidth]{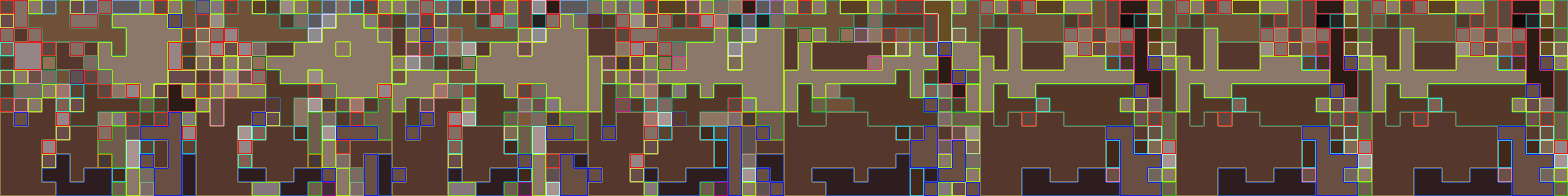}
  \end{subfigure}
  \begin{subfigure}{0.96\linewidth}
    \centering
    \includegraphics[width=1.0\linewidth]{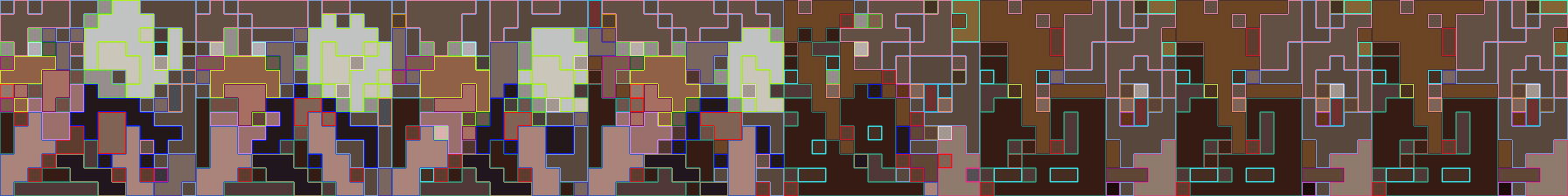}
  \end{subfigure}
  \caption{Visualisations of the difference in merging decisions made in layer $1$ versus layer $12$, produced with ViViT.}
  \label{fig:vivit_epickitchens_duplicate_layer_examples}
\end{figure*}

Secondly, in~\cref{fig:videomae_epickitchens_duplicate_layer_examples} and~\cref{fig:vivit_epickitchens_duplicate_layer_examples} we collect more examples of the differences in merging outcomes for the first and last layers of the models.
We note differences between how ViViT and VideoMAE merge tokens in the first layers, with VideoMAE creating large clusters with little consideration across frames, suggesting a spatial focus.
Notably, the tokens tend to be divided into an upper and lower half, which could possibly be due to the distribution of foreground and background in EK-$100$, where the top of the frame will usually portray kitchen background and the bottom of the frame will typically contain interacting objects.
On the other hand, ViViT tends to merge tokens across many frames within the first layer, yet these do not tend to occur around objects/distinguishable parts of the frame(s).
Comparatively, the behaviour for the final layers (across both models) displays the merging of tokens around objects in the scene and background.

\begin{figure*}[htp]
  \centering
  \begin{subfigure}{0.96\linewidth}
    \centering
    \includegraphics[width=1.0\linewidth]{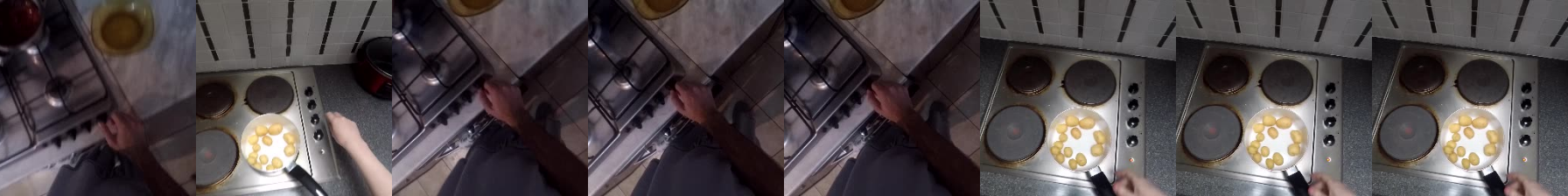}
  \end{subfigure}
  \begin{subfigure}{0.96\linewidth}
    \centering
    \includegraphics[width=1.0\linewidth]{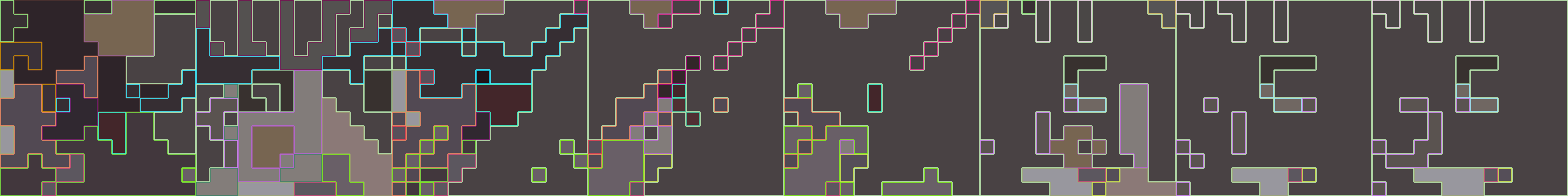}
  \end{subfigure}
  \hfill
  \begin{subfigure}{0.96\linewidth}
    \centering
    \includegraphics[width=1.0\linewidth]{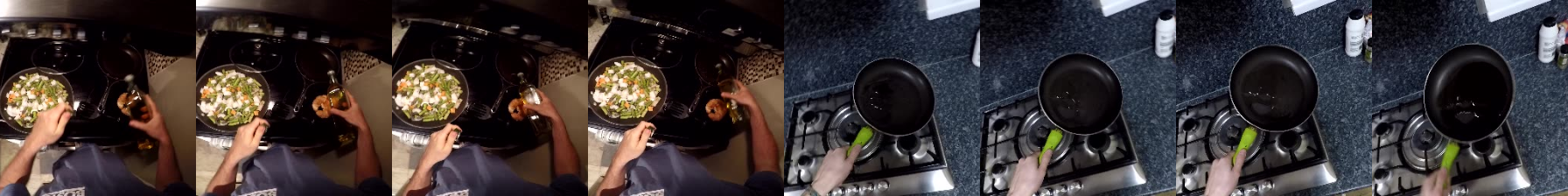}
  \end{subfigure}
  \begin{subfigure}{0.96\linewidth}
    \centering
    \includegraphics[width=1.0\linewidth]{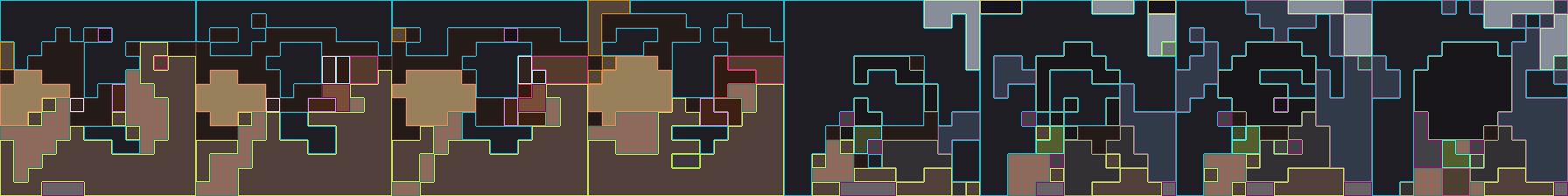}
  \end{subfigure}
  \hfill
  \begin{subfigure}{0.96\linewidth}
    \centering
    \includegraphics[width=1.0\linewidth]{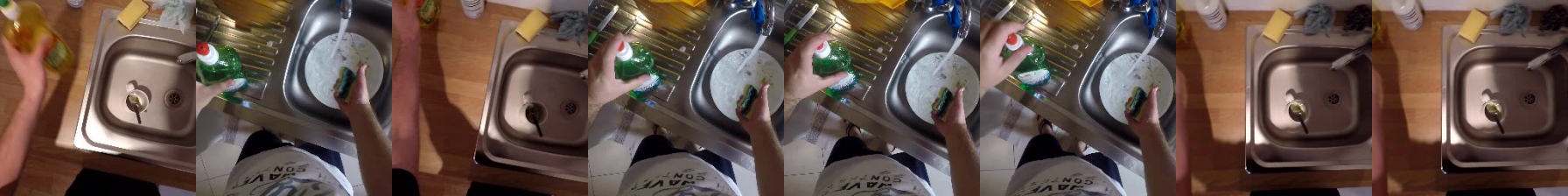}
  \end{subfigure}
  \begin{subfigure}{0.96\linewidth}
    \centering
    \includegraphics[width=1.0\linewidth]{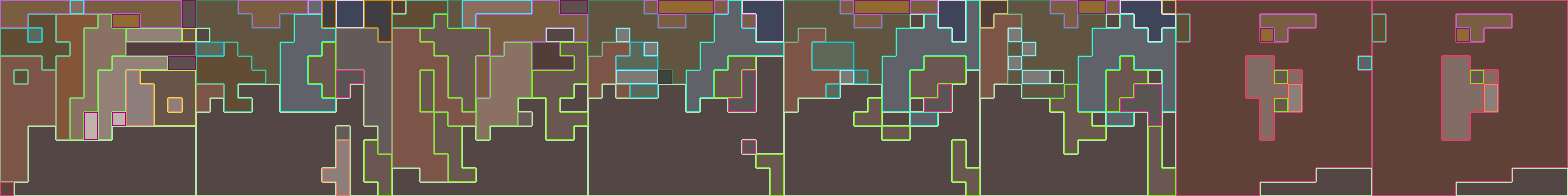}
  \end{subfigure}
  \hfill
  \begin{subfigure}{0.96\linewidth}
    \centering
    \includegraphics[width=1.0\linewidth]{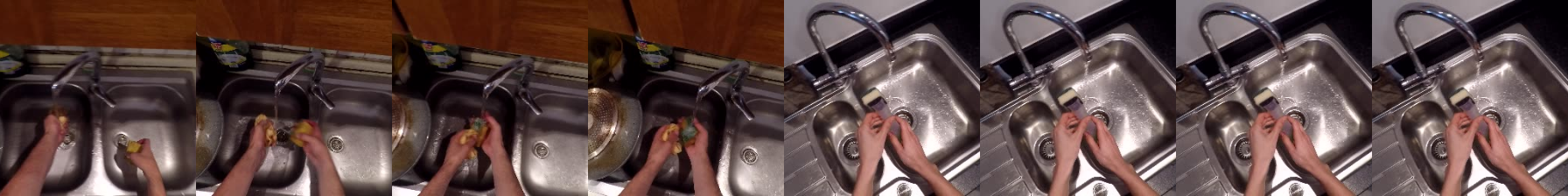}
  \end{subfigure}
  \begin{subfigure}{0.96\linewidth}
    \centering
    \includegraphics[width=1.0\linewidth]{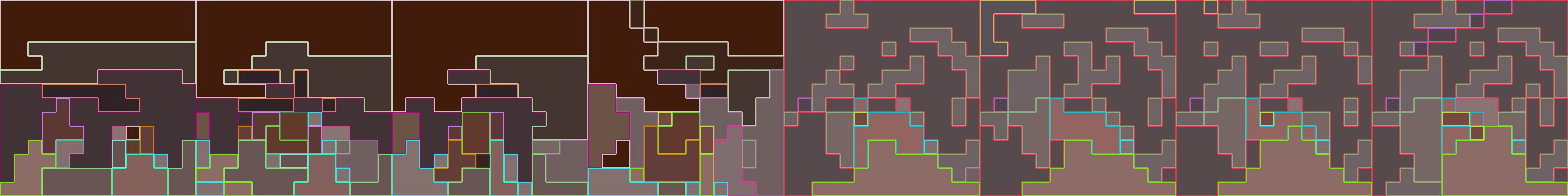}
  \end{subfigure}
  \hfill
  \begin{subfigure}{0.96\linewidth}
    \centering
    \includegraphics[width=1.0\linewidth]{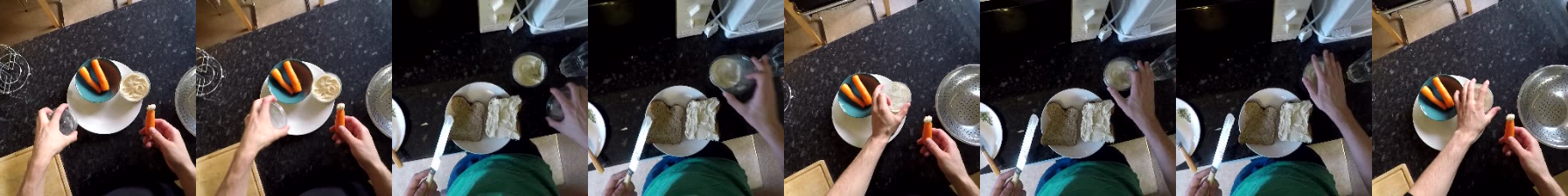}
  \end{subfigure}
  \begin{subfigure}{0.96\linewidth}
    \centering
    \includegraphics[width=1.0\linewidth]{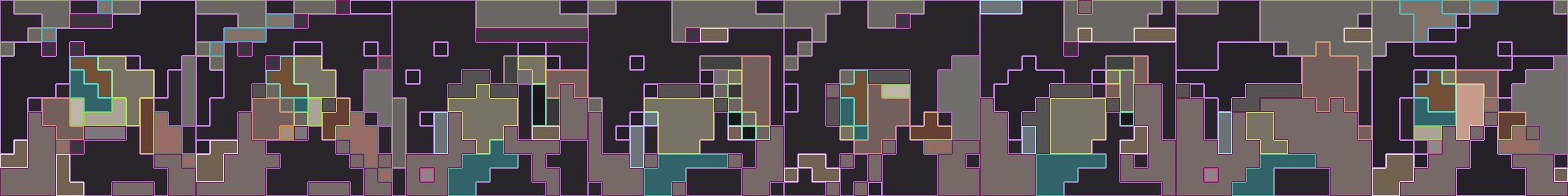}
  \end{subfigure}
  \caption{Merging outcomes for clips that have had half their frames from the most ``similar'' clip in the same noun class spliced in, produced with VideoMAE.}
  \label{fig:videomae_epickitchens_splicing_examples}
\end{figure*}

\begin{figure*}[htp]
  \centering
  \begin{subfigure}{0.96\linewidth}
    \centering
    \includegraphics[width=1.0\linewidth]{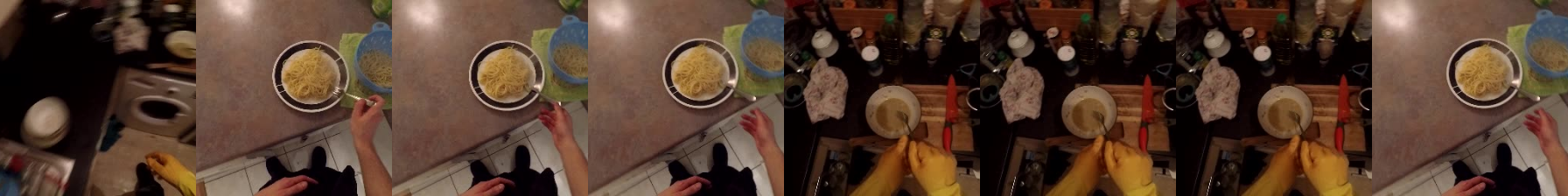}
  \end{subfigure}
  \begin{subfigure}{0.96\linewidth}
    \centering
    \includegraphics[width=1.0\linewidth]{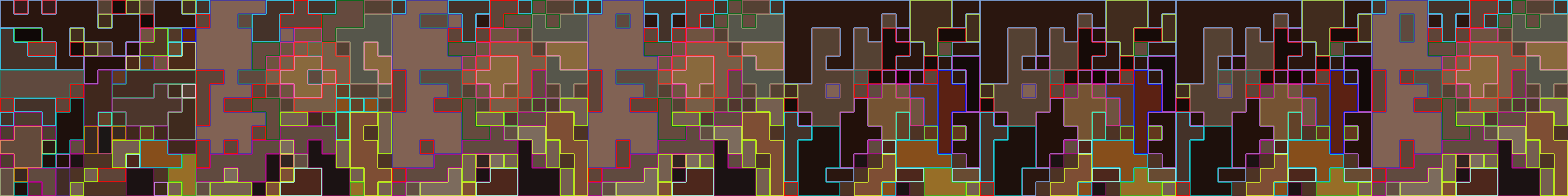}
  \end{subfigure}
  \hfill
  \begin{subfigure}{0.96\linewidth}
    \centering
    \includegraphics[width=1.0\linewidth]{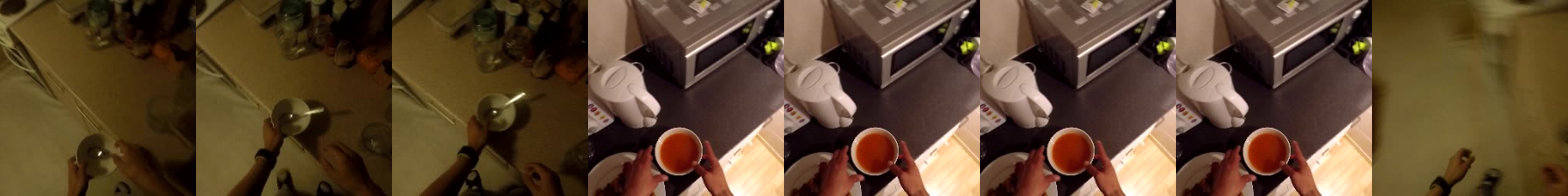}
  \end{subfigure}
  \begin{subfigure}{0.96\linewidth}
    \centering
    \includegraphics[width=1.0\linewidth]{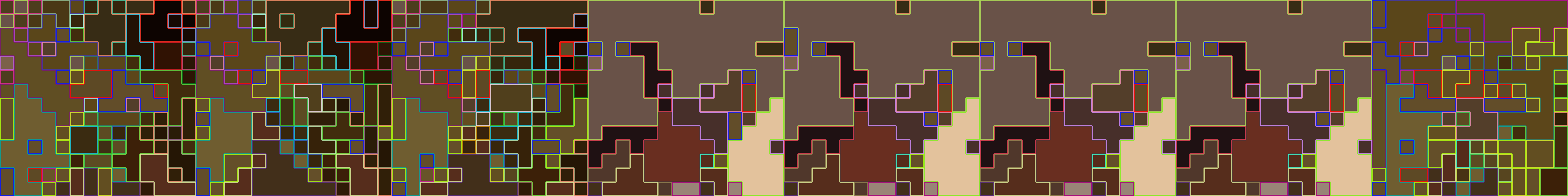}
  \end{subfigure}
  \hfill
  \begin{subfigure}{0.96\linewidth}
    \centering
    \includegraphics[width=1.0\linewidth]{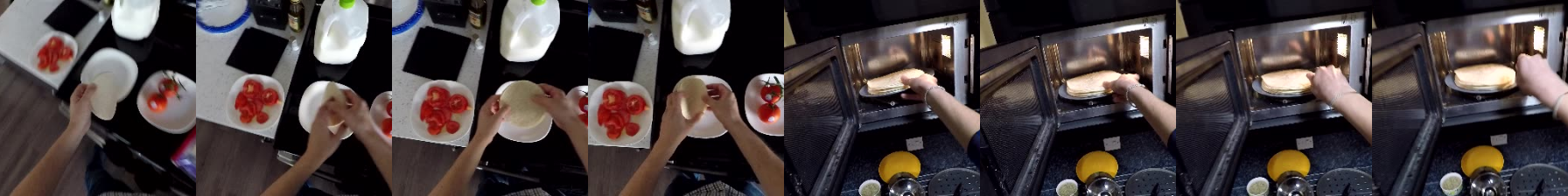}
  \end{subfigure}
  \begin{subfigure}{0.96\linewidth}
    \centering
    \includegraphics[width=1.0\linewidth]{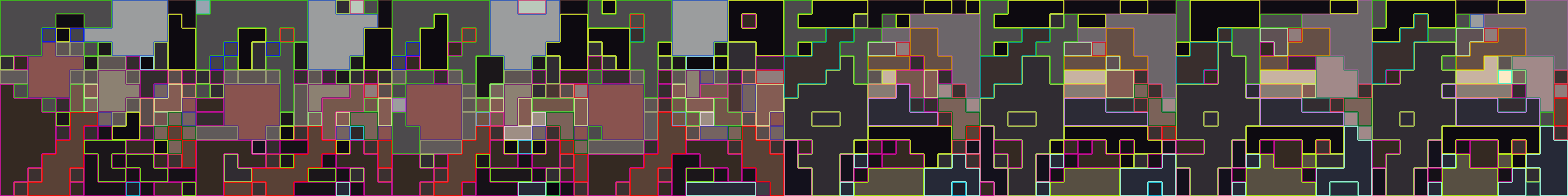}
  \end{subfigure}
  \hfill
  \begin{subfigure}{0.96\linewidth}
    \centering
    \includegraphics[width=1.0\linewidth]{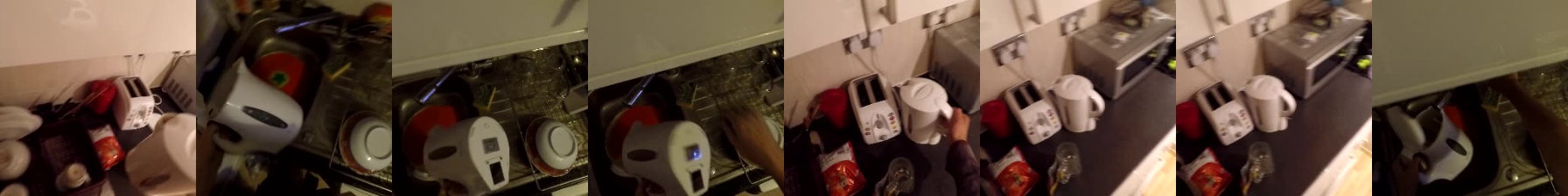}
  \end{subfigure}
  \begin{subfigure}{0.96\linewidth}
    \centering
    \includegraphics[width=1.0\linewidth]{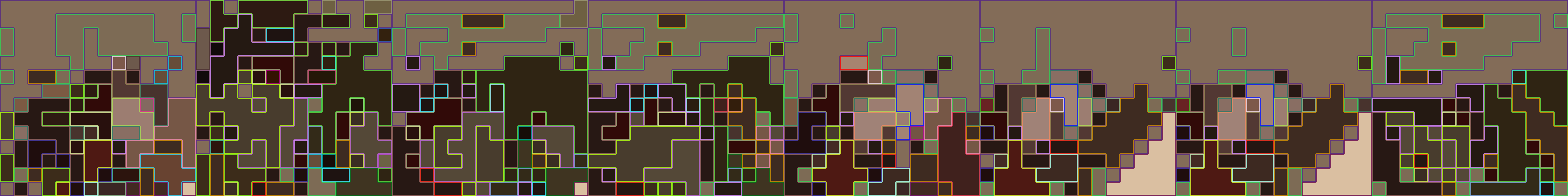}
  \end{subfigure}
  \hfill
  \begin{subfigure}{0.96\linewidth}
    \centering
    \includegraphics[width=1.0\linewidth]{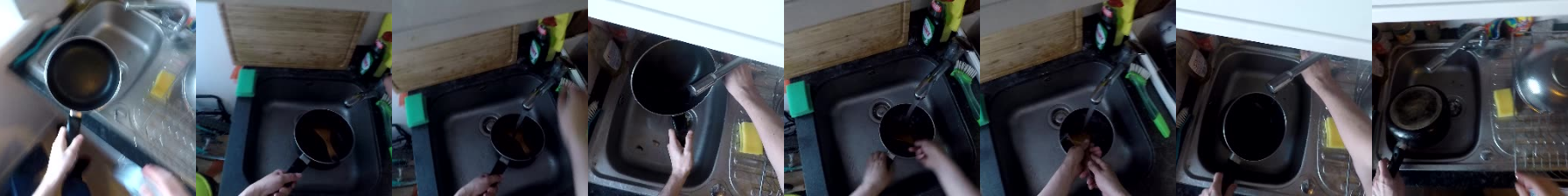}
  \end{subfigure}
  \begin{subfigure}{0.96\linewidth}
    \centering
    \includegraphics[width=1.0\linewidth]{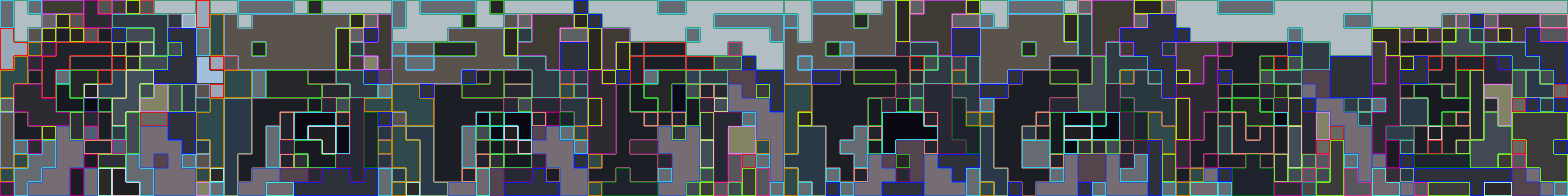}
  \end{subfigure}
  \caption{Merging outcomes for clips that have had half their frames from the most ``similar'' clip in the same noun class spliced in, produced with ViViT.}
  \label{fig:vivit_epickitchens_splicing_examples}
\end{figure*}

Finally, in~\cref{fig:videomae_epickitchens_splicing_examples} and~\cref{fig:vivit_epickitchens_splicing_examples} we generate more examples of clips where frames from the most ``similar'' clip in EK-$100$ have been spliced in, to demonstrate the lack of semantic merging.
In these cases, we have given the model the fairest chance by picking examples that also appear similar to the human eye.
Much of the examples appear to demonstrate that the model is only merging within either the original clip \textit{or} the spliced in clip, not between them.
There appear to be some examples for which the participants' hands are merged between the spliced frames, likely due to the fact that there are few possible visual differences for these tokens.
As discussed in the main paper, there is little to no evidence of the token merging process occurring between semantically relevant tokens, instead the token merging process is predominantly visual.

\clearpage

\end{document}